\pdfoutput=1

\documentclass[11pt]{article}

\usepackage[final]{coling}

\usepackage{times}
\usepackage{latexsym}

\usepackage[T1]{fontenc}

\usepackage[utf8]{inputenc}

\usepackage{microtype}

\usepackage{inconsolata}

\usepackage{graphicx}

\usepackage{enumitem}
\usepackage{colortbl}
\usepackage{bbm}
\usepackage{xspace}
\usepackage[normalem]{ulem}
\useunder{\uline}{\ul}{}
\usepackage{multirow}
\usepackage{listings}
\usepackage{amsmath}
\usepackage{amssymb}
\usepackage{mathtools}
\usepackage{amsthm}
\usepackage{graphicx}
\usepackage{wrapfig}
\usepackage{subfigure}
\usepackage{hyperref}
\usepackage{makecell}
\hypersetup{
    colorlinks=true,
    citecolor=blue,
    linkcolor=blue,
}
\newcommand{\ie}{\emph{i.e.,}\xspace}
\newcommand{\eg}{\emph{e.g.,}\xspace}

\definecolor{codegray}{rgb}{0.5,0.5,0.5}
\definecolor{codepurple}{rgb}{0.58,0,0.82}
\definecolor{backcolour}{rgb}{0.95,0.95,0.92}
\lstdefinestyle{mystyle}{
    backgroundcolor=\color{backcolour},   
    numberstyle=\tiny\color{codegray},
    basicstyle=\ttfamily\footnotesize,
    stringstyle=\color{codepurple},
    breakatwhitespace=false,         
    breaklines=true,                 
    captionpos=b,                    
    keepspaces=true,             
    numbers=right,                    
    numbersep=5pt,                  
    showspaces=false,                
    showstringspaces=false,
    showtabs=false,                  
    tabsize=2,
    breakautoindent=true, 
    breakindent=0pt
}
\newcommand{\colorr}[1]{\textcolor{black}{#1}}
\newcommand{\colorb}[1]{\textcolor{blue}{#1}}

\newcommand{\zc}[1]{\textcolor{black}{#1}}
\newcommand{\xd}[1]{\textcolor{black}{#1}}
\newcommand{\gxd}[1]{\textcolor{black}{#1}}
\newtheorem{theorem}{Theorem}[section]
\newtheorem{proposition}[theorem]{Proposition}

\theoremstyle{definition}
\newtheorem{definition}[theorem]{Definition}

\theoremstyle{remark}

\title{\zc{Planning with Multi-Constraints via Collaborative Language Agents}}

\author{
 \textbf{Cong Zhang\textsuperscript{*}},
 \textbf{Derrick Goh Xin Deik\textsuperscript{*}},
 \textbf{Dexun Li\textsuperscript{}},
 \textbf{Hao Zhang\textsuperscript{}},
 and \textbf{Yong Liu\textsuperscript{}}
\\
 \textsuperscript{}Huawei Noah's Ark Lab
 \\
\texttt{cong.zhang92@gmail.com, goh.xin.deik@huawei.com, lidexun@huawei.com,} \\
\texttt{zhang.hao3@huawei.com, liu.yong6@huawei.com}
}

\begin{document}
\maketitle

\def\thefootnote{*}\footnotetext{These authors contributed equally to this work.}

\begin{abstract}
The rapid advancement of neural language models has sparked a new surge of intelligent agent research. Unlike traditional agents, large language model-based agents (LLM agents) have emerged as a promising paradigm for achieving artificial general intelligence (AGI) due to their superior reasoning and generalization capabilities. Effective planning is crucial for the success of LLM agents in real-world tasks, making it a highly pursued topic in the community. Current planning methods typically translate tasks into executable action sequences. However, determining a feasible or optimal sequence for complex tasks \zc{with multiple constraints} at fine granularity, which often requires compositing long chains of heterogeneous actions, remains challenging. This paper introduces \zc{Planning with Multi-Constraints (PMC)}, a \textbf{zero-shot} methodology for collaborative LLM-based multi-agent systems that simplifies complex task planning \zc{with constraints} by decomposing it into a hierarchy of subordinate tasks. Each \zc{subtask} is then mapped into executable actions. \zc{PMC} was assessed on two \zc{constraint-intensive} benchmarks, TravelPlanner and API-Bank. Notably, \zc{PMC} achieved an average \zc{$42.68\%$} success rate on TravelPlanner, significantly higher than \zc{GPT-4} ($2.92\%$), and outperforming
\zc{GPT-4}
with ReAct on API-Bank by \zc{$13.64\%$}, showing the immense potential of integrating LLM with multi-agent systems. \zc{We also show that PMC works with small LLM as the planning core, e.g., \xd{LLaMA-3.1-8B}.} Our code is publically available at https://github.com/zcaicaros/PMC.
\end{abstract}

\section{Introduction}
\zc{
Recently, there has been a growing interest in using large language models (LLMs) as the cognitive core of agents~\cite{wang2024survey}, due to their ability to understand and execute human instructions in natural language. LLM-powered agents, known for their strong logical skills and strategic planning, are considered a promising path toward achieving artificial general intelligence (AGI)~\cite{wang2024survey, you2024far}.}

Current LLM agent planning solutions aim to map tasks to sequences of executable actions~\cite{huang2024understanding}. The \textit{plan-then-execute} methods~\cite{shen2024hugging,wang2023plan} break down complex tasks into small, manageable sub-tasks to facilitate the inference of a sequence of executable actions. In contrast, the \textit{step-by-step} methods~\cite{wei2022chain,yao2023react,chen2023program,wu2023visual,gao2023pal} interleave planning and execution, where each action is determined based on previous outcomes. The former simply assumes each sub-task can be executed with a single tool~\cite{shen2024hugging}, but real-world applications often require tools with diverse functionalities~\cite{krishnakumar1995managing}. The latter is unsuitable for time-sensitive constraints requiring comprehensive condition assessment, meticulous planning and subsequent execution. Moreover, the piecemeal nature of these approaches may lead to suboptimal outcomes and potential task failure. To improve planning stability and performance, recent studies~\cite{dagan2023dynamic,guan2023leveraging,yang2023coupling} integrate LLMs with external planning tools requiring task descriptions in specific formats, \eg first-order logic~\cite{barwise1977introduction}. However, translating various tasks into certain computational logic can be challenging and often demands a myriad of domain knowledge~\cite{cropper2022inductive}. Existing LLM-based multi-agent systems primarily simulate human behaviors and social activities~\cite{li2024camel,park2023generative,gao2023large}, while planning for collaborative multi-agent systems \zc{under multiple constraints}, despite their significant potential, remains underexplored.

In this work, we propose \zc{\textbf{P}lanning with \textbf{M}ulti-\textbf{C}onstraints} (\zc{PMC}), a \textit{zero-shot} planning method for collaborative LLM-based multi-agent systems. \zc{PMC} simplifies complex task planning by breaking it down into a hierarchy of subordinate tasks, each achievable through a series of (heterogeneous) tool calls. Specifically, \zc{PMC} comprises a \textit{manager} agent for task decomposition and a fleet of \textit{executor} agents to perform \zc{sub-task}s. The manager performs \textit{task-level planning} by decomposing the task into a graph, where each node represents a specific \zc{sub-task (e.g., recommendation)} and the edges delineate the dependency topology among tasks. Then each \zc{sub-task} is decomposed into a sequence of function calls, \ie \textit{step-level planning and execute}, via an executor. The executor may utilize off-the-shelf planning techniques, like ReAct~\cite{yao2023react}, to facilitate \zc{sub-task} accomplishment. \zc{PMC} can be viewed as a framework that extends the capabilities of individual LLM agents by equipping with \zc{cooperation} cores, thus transforming them into \zc{collaborative multi-agent system. PMC focuses on complex planning scenarios with constraints, such as budget limitations, which are categorized into "local" and "global" types.} Local constraints are managed by executors during \zc{sub-task} execution, whereas global constraints are considered in conjunction with other variables. To improve the success rate and stability, \zc{PMC} employs a \textit{supervisor} agent to \gxd{refine a sub-task if the results of the previous sub-tasks it depends on are obtained} and a \textit{deliverer} agent to produce the final outcome. \gxd{If the given sub-task does not depend on any previous sub-task, the supervisor agent will not refine the sub-task. The deliverer agent will either summarize the outcome or make a decision for the user based on the result of all sub-tasks.}

Distinct from the toy tasks~\cite{singh2023progprompt} or puzzles~\cite{ahn2024large} commonly used in existing planning methods, we evaluate \zc{PMC} on two real-world applications: itinerary planning and daily tool usage. Experiment results show that \zc{PMC} achieves substantial performance gains on two benchmarks. Specifically, \zc{PMC} obtains \zc{\textbf{$42.68\%$}} success rate on TravelPlanner~\cite{xie2024travelplanner}, a significant increase from \zc{GPT-4 (\textbf{$2.92\%$})}. It also surpasses \zc{GPT-4} with ReAct on API-Bank~\cite{li2023api} by \zc{$13.62\%$} in absolute improvement. To the best of our knowledge, \zc{PMC} is the first plan-and-execute method for collaborative LLM-based multi-agent systems \zc{to effectively address complex tasks involving multiple constraints}. \zc{Moreover, Ww show that \xd{LLaMA-3.1-8B} equipped with PMC under one demonstration example surpasses GPT-4 by a large margin.}

\section{Literature Review for Language Model Agent Planning}

The emergence of LLMs introduces new paradigms for agents~\cite{chu2023cotsurvey,wang2024autonomous,masterman2024landscape}, demonstrating significant intelligence in reasoning~\cite{kojima2022large,wei2023chainofthought,wang2023selfconsistency}, planning~\cite{yao2023react,yao2023tree,besta2024graph}, instruction-following~\cite{xu2023wizardlm,wang2023selfinstruct,ren2023selfevaluation}, and tool-usage~\cite{schick2023toolformer,yang2023tools,shen2024hugging} across various domains. Planning acts as an essential capability to interact with external environments, which involves organizing thought trajectories, setting objectives, and determining steps to accomplish the objectives~\cite{mattar2022planning}. Some work~\cite{wei2023chainofthought,yao2023react,chen2023program,wang2023plan} focuses on task decomposition, aiming to solve complex tasks in a divide-and-conquer manner. The plan selection methods~\cite{yao2023tree,besta2024graph,wang2023selfconsistency,xiao2023llm} elicit LLMs to generate various alternative plans for a task following by a search algorithm for optimal plan selection and execution. Recent studies~\cite{shinn2023reflexion,madaan2023selfrefine,huang2022inner,gou2024critic} also explore to enhance LLM's planning ability via reflection and refinement strategies. Moreover, some work~\cite{liu2023llmp,lin2023swiftsage,zhao2024expel} also introduces external planners to aid the planning procedure of LLMs.

Numerous strategies have been developed to harness the potentials of LLMs for specific agent planning~\cite{xi2023rise}, whose effectiveness and accuracy of planning significantly determine the agent's robustness and usability. Web-agents~\cite{yao2022webshop,deng2023mind2web,gur2024webagent,furuta2024exposing} explore the interaction between LLM and web-environment by simulating human's web-browsing behaviors via RL-based planning or trajectory planning. General tool-agents require to interact with massive APIs or tools, making the planning procedures more challenging. Solutions to tool-agent planning usually rely on various task decomposition~\cite{yuan2024easytool,shen2024hugging}, self-rectification~\cite{ma2024sciagent} and domain-reasoning~\cite{lu2023chameleon} strategies. Other task-specific agents focus on designing sophisticated planning strategies, such as tree search~\cite{zhou2023language} and Bayesian adaptive MDPs~\cite{liu2023reason}. Multi-agent systems~\cite{chen2024agentverse,hong2024meta,gong2023mindagent,mei2024aios} seek to solve more complex real-world tasks by combining multiple powerful LLM-based agents. Existing solutions mainly focus on tackling the complexities inherent in integrating heterogeneous agents with different capabilities and specializations~\cite{mei2024aios}, while the planning strategies among these agents are overlooked. In contrast, our \zc{PMC} focuses on designing generalized, robust planning strategies for multi-agent systems. Although $\texttt{LLM}_{api}$Swarm~\cite{zhuge2024language} shares a similar concept with \zc{PMC}, it focuses on visualizing multi-agent collaboration via composite graphs to aid prompt tuning, while our \zc{PMC} is a planning algorithm specifically designed for systems with multiple collaborative agents.

\section{Preliminaries}


\zc{The LLM-based agent is an AI system utilizing an LLM as its computational core, enabling functionalities beyond text generation such as task execution, logical reasoning, and autonomous operation. Formally, an LLM agent includes: $A=(\mathtt{LLM}, \mathcal{F}_n, R, \mathcal{S}, C)$. $\mathtt{LLM}$ is the language model instance (e.g., LLaMA~\cite{touvron2023llama}) used for reasoning, planning, and decision-making. $\mathcal{F}_n$ is a set of functions/actions performed by the agent. $R$ is the agent's role as defined by the prompt. $S \in \mathcal{S}$ represents the agent's dynamic state, including knowledge and internal processes. $C$ is the communication module for exchanging information with other agents or the environment. In a collaborative LLM-based multi-agent system, multiple agents, $[A_1, A_2, \dots, A_m]$, work together to achieve a common goal. Each agent $A_i$ has a specific role $R_i$ and task specialization, \eg task decomposition.}

\section{\zc{Planning with Multi-Constraints}}
\begin{figure*}[!t]
    \begin{center}    
    \centerline{\includegraphics[width=1\textwidth]{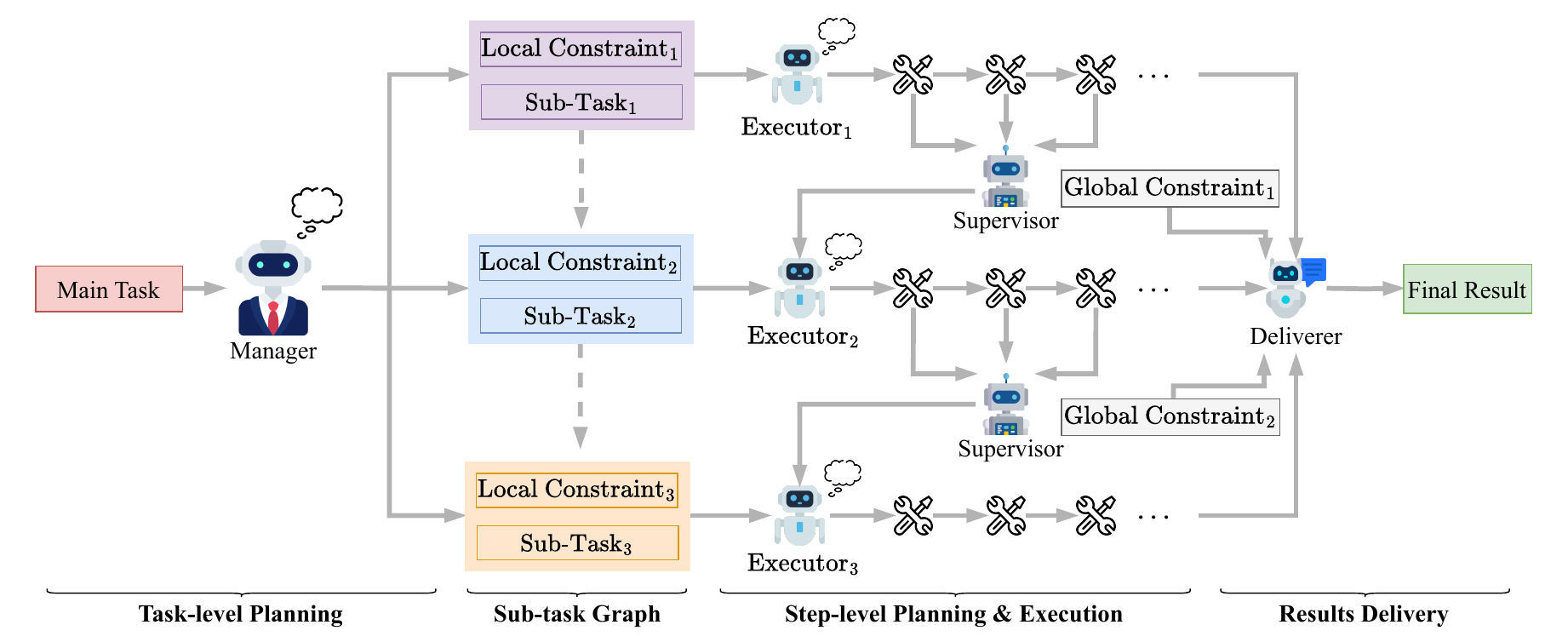}}
    \caption{An overview of \zc{PMC} Framework. The \zc{PMC} Framework provides a structured methodology for managing and executing \zc{sub-task}s within a directed \zc{sub-task} graph topology, as the manager coordinates. For instance, the completion of $\text{Task}_2$ depends on the outputs derived from $\text{Task}_1$, which a supervisor agent subsequently consolidates. The executor agent is tasked with implementing the \zc{sub-task}, considering any local constraints present. Upon completion of the \zc{sub-task}s, the deliverer agent is responsible for aggregating all \zc{sub-task} outcomes to satisfy global constraints and subsequently achieve the overarching task objectives.}
    \label{fig:framework}
    \end{center}
    \vskip -0.2in
\end{figure*}

Complex projects, such as those in construction or manufacturing, often present significant challenges, particularly with diverse, geographically dispersed teams. However, thorough planning, effective communication, and collaboration can ensure successful outcomes~\cite{lester2017planningblocks}. In light of this, we introduce \zc{planning with multi-constraints} (\zc{PMC}), a novel zero-shot planning approach for collaborative LLM-based multi-agent systems to \zc{address complex tasks with multiple constraints}. In \zc{PMC}, a designated \textit{manager} agent decomposes a complex task $\mathcal{T}=\{T_i | i \in S(K) = \{1, \cdots, K\}\}$ into smaller, more manageable sub-tasks, $\{T_1, T_2, T_3, \cdots, T_K\}$ through \textit{task-level planning}. These \zc{sub-task}s are then converted into a sequence of heterogeneous tool-using actions executed by a fleet of \textit{executor} agents, a process referred to as \textit{step-level planning and execution}. Additionally, \zc{PMC} incorporates a supervisor agent to facilitate sharing synthesized \zc{sub-task} outcomes among executors and a deliverer agent to consolidate final results upon the collective findings of all \zc{sub-task}s. 
The comprehensive framework of \zc{PMC} is illustrated in Figure~\ref{fig:framework}.

\subsection{The Collaborative Multi-Agent System Design}
\begin{figure*}[!ht]
    \centering
    \subfigure[Manager.]{\includegraphics[width=.245\textwidth]{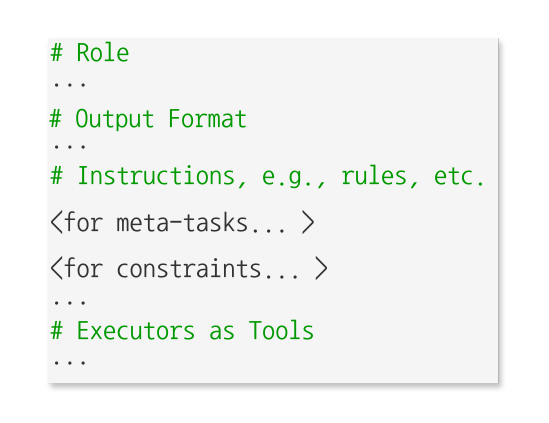}\label{fig:agent-design-manager}}
    \subfigure[Executors.]{\includegraphics[width=.245\textwidth]{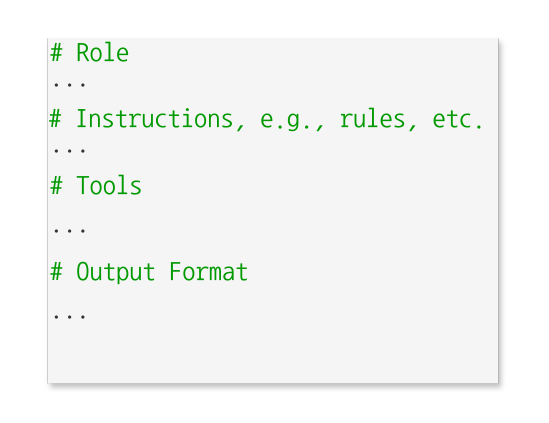}\label{fig:agent-design-executor}}
    \subfigure[Supervisor.]{\includegraphics[width=.245\textwidth]{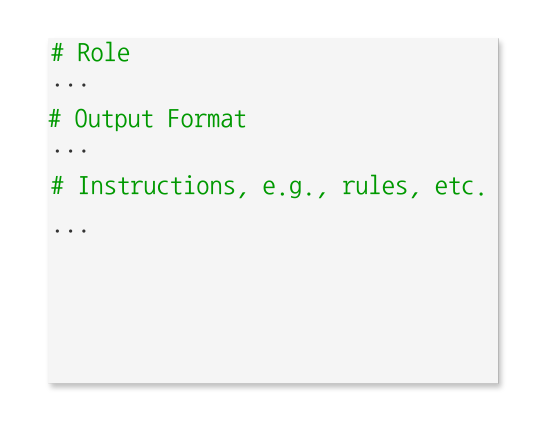}\label{fig:agent-design-supervisor}}
    \subfigure[Deliverer.]{\includegraphics[width=.245\textwidth]{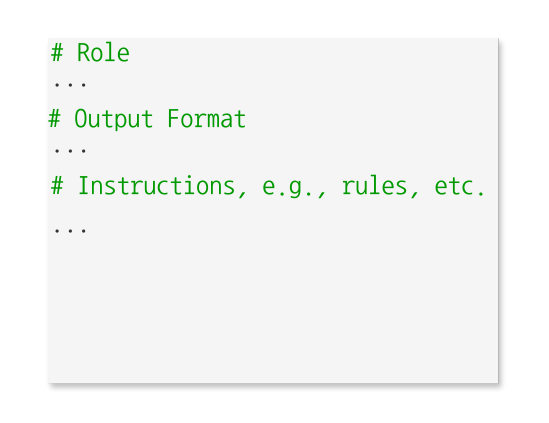}\label{fig:agent-design-deliverer}}
    \caption{The zero-shot prompt structure for each agent. Note both supervisor and deliverer agents do not require function calls, while different executors will have different tool lists.}
    \label{fig:agent-design}
\end{figure*}

\gxd{The description of each agent is shown in each subsection and the prompt structure of each agent is illustrated in Figure ~\ref{fig:agent-design}.}

\subsubsection{Manager Agent}
The manager agent has two primary objectives. Firstly, it decomposes the intricate task $\mathcal{T}$ into a set of interconnected \zc{sub-task}s $\mathcal{T}=\{T_i | i \in S(K) = \{1, \cdots, K\}\}$. These \zc{sub-task}s often exhibit dependencies, where completing one task is contingent on completing another. For instance, deciding on hotels usually depends on finalizing the trip destination. Thus, the manager must identify and define these inter-dependencies meticulously. Additionally, the manager has to assign suitable executors to each \zc{sub-task}. Executors are viewed as a collection of composite tools, and the manager matches them based on the \zc{sub-task} requirements, a method termed the \textit{executor as tools} technique.

Secondly, the manager must make well-informed decisions on task assignments, especially under constraints like budget limits or specific transportation needs in travel scenarios. Some constraints can be managed during individual \zc{sub-task}s. For instance, for a \zc{sub-task} that searching for accommodation, the minimum stay requirement ensures only suitable hotels are considered. The manager also needs to identify constraints that interact with other variables across \zc{sub-task}s and cannot be solved within a single \zc{sub-task}. For instance, when selecting a flight, available hotel and restaurant options provided by other \zc{sub-task}s must be jointly considered. Identifying local and global constraints to \textit{divide and conquer} them is crucial for successfully completing complex tasks.
The formal definitions for the local and global constraints are presented as follows:

\begin{definition}
    A constraint $C_l$ is \textit{local} if and only if $\exists!\ i \in S(K)$ such that $C_l$ can be fulfilled purely based on the results of $T_i$.
    \label{def:local-constraint}
\end{definition}
    
\begin{definition}
    A constraint $C_g$ is \textit{global} if and only if $\exists\ \mathcal{T}_{C_g}=\{T_i|i\in S(K)\} \subseteq S(K)$ and $K \geq |\mathcal{T}_c| >1$, such that $C_g$ can be fulfilled based on the results of all $T_i \in \mathcal{T}_{C_g}$, where $|\cdot|$ demotes the cardinality of a set.
    \label{def:global-constraint}
\end{definition}

It is important to note that the manager identifies potential constraints and categorizes them into local and global ones exclusively based on internal knowledge. No prior information about the constraints for $\mathcal{T}$ is provided, ensuring that the zero-shot property of \zc{PMC} is maintained. Figure~\ref{fig:agent-design-manager} depicts the logic for manager prompt design. 

\begin{figure*}[t]
    \begin{center}    \centerline{\includegraphics[width=1\textwidth]{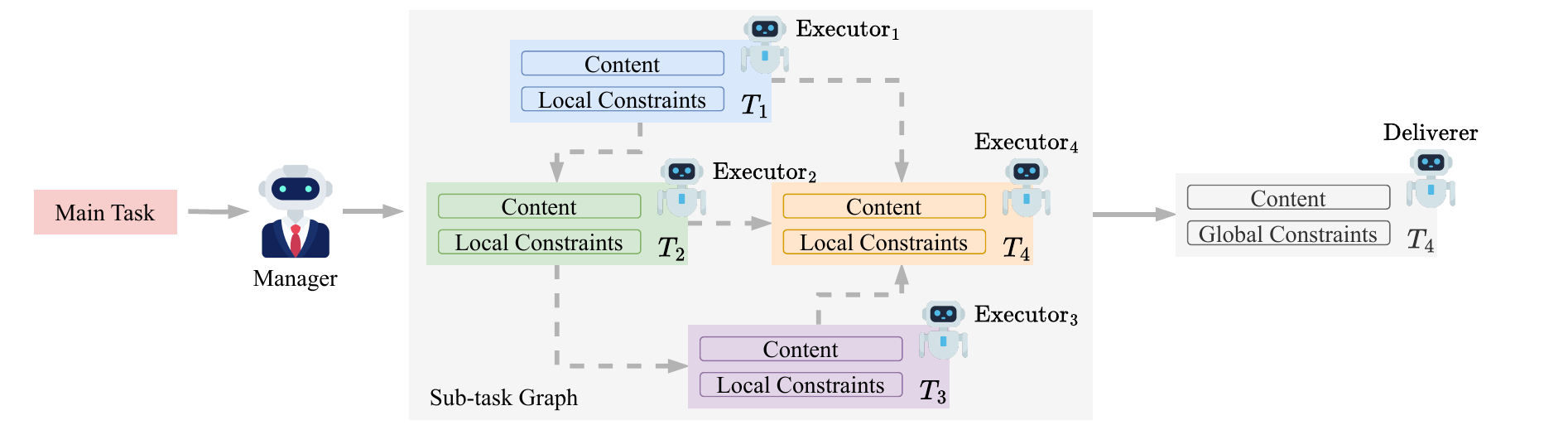}}
    \caption{An overview of \zc{sub-task} graph, which reveals the task-level decomposition. The manager agent decomposes the main task into several \zc{sub-task}s with inter-dependencies (dashed arrows).}
    \label{fig:meta-task-graph}
    \end{center}
    \vskip -0.2in
\end{figure*}

\subsubsection{Executor Agent}
The manager agent assigns each \zc{sub-task} to an executor agent, which has access to various heterogeneous tools (\eg functions). The executor aims to create a sequence of actions (\eg function calls) to complete the assigned \zc{sub-task} while adhering to local constraint $C_l$. This process significantly reduces the planning complexity, as the executor focuses on a specific, well-defined task with clear requirements and constraints. Consequently, applying off-the-shelf single-agent planning methods to map a task to an execution sequence is feasible. Figure~\ref{fig:agent-design-executor} illustrates the conceptual prompt design of the executor. Owing to the functional variation among executors, a tailored design approach is necessary for each executor, depending on the specific tools available to them.

\subsubsection{Supervisor Agent}
The role of the supervisor agent is to refine the \zc{sub-task} $T_i$ by incorporating synthesized outcomes from neighboring \zc{sub-task}s. After the manager agent decomposes the main task, only ambiguous objectives (\eg ``\textit{Finding a hotel in city B}'') and inter-dependencies among \zc{sub-task}s (\eg ``\textit{Searching flight to New York}'' $\rightarrow$ ``\textit{Finding a hotel in city B}'') are identified. To execute  $T_i$ effectively, its input parameters need precise specifications. For instance, the input ``\textit{Finding a hotel in city B}'' must be correctly \zc{instantiated as} ``\textit{Finding a hotel in New York near John F. Kennedy International Airport}'' based on outcomes (\eg ``\textit{Booked flight ZC9896 to New York, arriving at John F. Kennedy International Airport}'') from preceding \zc{sub-task} (\eg ``\textit{Searching flight to New York}''). To address these nuances and eliminate ambiguities, the supervisor agent acts before the commencement of $T_i$. It rewrites $T_i$ by referencing the outcomes of all neighboring \zc{sub-task}s. This ensures all necessary parameters are included and accurately instantiated. Here, the neighbor of $T_i$ is defined as the collection of \zc{sub-task}s that have direct inter-dependencies with $T_i$. Formally, the neighborhood of $T_i$ is defined as:

\begin{definition}
    The neighbors $\mathcal{N}(T_i)$ of \zc{sub-task} $T_i$ is defined as $\{T_j|\forall j \in S(K), s.t., T_j \rightarrow T_i\}$.
    \label{def:neighborhood-of-meta-task-i}
\end{definition}

\textbf{\textit{Remark}}: An alternative idea is to include all precedent \zc{sub-task}s in outcome synthesis for $T_i$ instead of just its immediate neighbors. While this seems reasonable, it can overwhelm the manager agent, especially given the input limitations of LLMs with many tasks. In contrast, our proposed approach focuses on immediate neighbors, maintaining manageability and avoiding such complexities. Moreover, our approach is capable of preserving all information through the ``message-passing'' mechanism, allowing correct results from precedent \zc{sub-task}s to be recursively propagated to $T_i$. The following proposition supports this assertion:

\begin{proposition}
    \textit{A \zc{sub-task} $T_i$ is accomplishable while adhering to local constraints if and only if all the \zc{sub-task}s within its direct neighborhood $\mathcal{N}(T_i)$ are accomplished with their respective local constraints maintained.}
\label{proposition: 1}
\end{proposition}

The proof is \zc{in Appendix~\ref{proof: proposition 1}}.
The supervisor prompt design is delineated in Figure~\ref{fig:agent-design-supervisor}.

\begin{figure*}[t]
    \begin{center}
    \centerline{\includegraphics[width=0.8\textwidth]{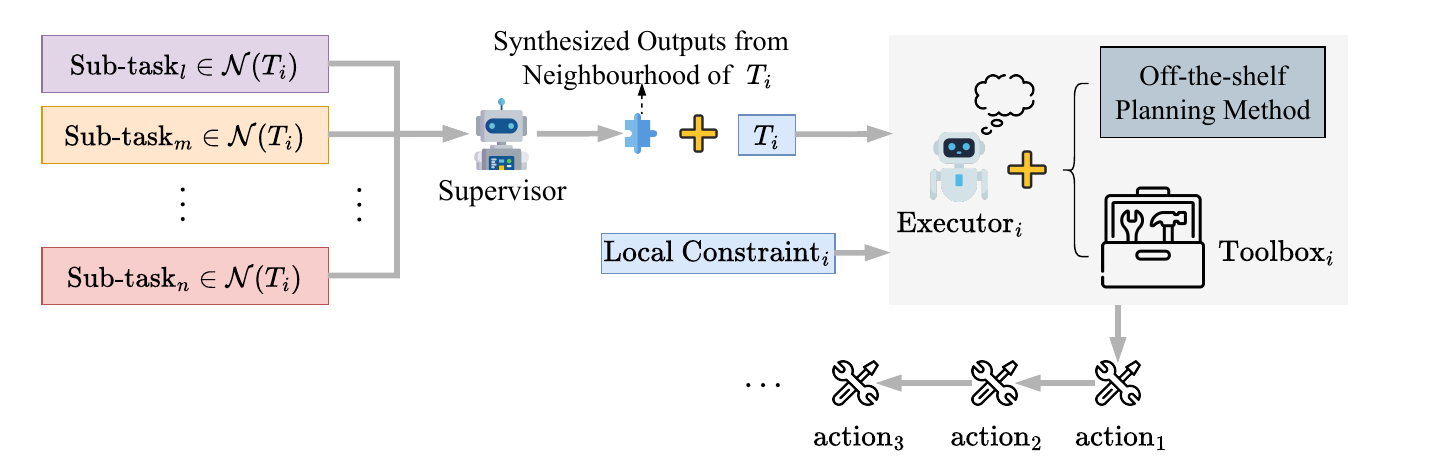}}
    \caption{step-level Planning and Execution. The executor is furnished with a planning core and a toolbox comprising diverse functions. This includes an off-the-shelf planning algorithm such as ReAct~\cite{yao2023react}, which is used to translate the \zc{sub-task} into a series of executable function calls required to accomplish the assigned \zc{sub-task}.}
    \label{fig:step-level-planning-and-execution}
    \end{center}
    \vskip -0.2in
\end{figure*}

\subsubsection{Deliverer Agent}
The primary objective of the deliverer agent is to synthesize the outcomes of all \zc{sub-task}s while ensuring alignment with the global constraints, $\{C_g\}$. This synthesis is critical because $\{C_g\}$ can only be effectively addressed once all \zc{sub-task} results are available. Thus, the deliverer agent is uniquely positioned to manage these constraints, ensuring that the final results comprehensively satisfy all global constraints. The logic of the deliverer prompt design is depicted in Figure~\ref{fig:agent-design-deliverer}.

The detailed prompt design and technical introduction of all agents are presented in Appendix~\ref{appendix:prompt-design}. \colorr{We provide the process of constructing and optimising the prompts in Appendix~\ref{appendix:prompt-optimization}. We hope this prompt construction process will give the community informative suggestions on how to build complex prompts from scratch.}


\begin{table*}[h]
\centering
\caption{The Average Pass Rates (\%) ``\textbf{With} Unconventional Hint'' for Instances Across All Difficulty Level \zc{on TravelPlanner}. The highest final pass rates are highlighted in bold blue.}
\label{tab:overall-performance-TravelPlanner-with-commensense}
\setlength{\extrarowheight}{3pt}
\scalebox{0.6}{
\begin{tabular}{@{}ccc|cc|cc|c@{}}
\hline
\Xhline{1.5pt}
\multicolumn{3}{c|}{With Unconventional Hint}                             & \zc{GPT-3.5} + ReAct          & \zc{GPT-4} + ReAct          & \zc{GPT-3.5} + \zc{PMC}            & \zc{GPT-4} + \zc{PMC}            & \zc{GPT-4} (SP)             \\ 
\Xhline{1pt}
\multicolumn{1}{c|}{}           & \multicolumn{2}{c|}{Delivery Rate}      &98.33 \cellcolor[HTML]{EFEFEF} &98.33 \cellcolor[HTML]{EFEFEF} &91.67 \cellcolor[HTML]{EFEFEF} &96.67 \cellcolor[HTML]{EFEFEF} &100.00 \cellcolor[HTML]{EFEFEF} \\ 
\Xcline{2-3}{0.4pt}
\multicolumn{1}{c|}{}           & \multicolumn{1}{c|}{Common-}    & Micro &74.38 \cellcolor[HTML]{EFEFEF} &79.38 \cellcolor[HTML]{EFEFEF} &63.54 \cellcolor[HTML]{EFEFEF} &87.29 \cellcolor[HTML]{EFEFEF} &92.08 \cellcolor[HTML]{EFEFEF} \\ 
\Xcline{3-3}{0.4pt}
\multicolumn{1}{c|}{Validation} & \multicolumn{1}{c|}{sense}      & Macro &0.00 \cellcolor[HTML]{EFEFEF} &8.33 \cellcolor[HTML]{EFEFEF} &1.67 \cellcolor[HTML]{EFEFEF} &43.33 \cellcolor[HTML]{EFEFEF} &50.00 \cellcolor[HTML]{EFEFEF} \\ 
\Xcline{2-3}{0.4pt}
\multicolumn{1}{c|}{Set (60)}   & \multicolumn{1}{c|}{Hard}       & Micro &0.71 \cellcolor[HTML]{EFEFEF} &7.14 \cellcolor[HTML]{EFEFEF} &0.71 \cellcolor[HTML]{EFEFEF} &47.14 \cellcolor[HTML]{EFEFEF} &52.86 \cellcolor[HTML]{EFEFEF} \\ 
\Xcline{3-3}{0.4pt}
\multicolumn{1}{c|}{}           & \multicolumn{1}{c|}{Constraint} & Macro &0.00 \cellcolor[HTML]{EFEFEF} &5.00 \cellcolor[HTML]{EFEFEF} &1.67 \cellcolor[HTML]{EFEFEF} &46.67 \cellcolor[HTML]{EFEFEF} &28.33 \cellcolor[HTML]{EFEFEF} \\ 
\Xcline{2-3}{0.4pt}
\multicolumn{1}{c|}{}           & \multicolumn{2}{c|}{\colorr{Final Pass Rate}}    &0.00 \cellcolor[HTML]{EFEFEF} &1.67 \cellcolor[HTML]{EFEFEF} &0.00 \cellcolor[HTML]{EFEFEF} &\textbf{\colorb{33.33}} \cellcolor[HTML]{EFEFEF} &13.33 \cellcolor[HTML]{EFEFEF} \\ 
\Xhline{1.5pt}
\hline
\multicolumn{1}{c|}{}           & \multicolumn{2}{c|}{Delivery Rate}      &93.50 \cellcolor[HTML]{DAE8FC} &98.38 \cellcolor[HTML]{DAE8FC} &84.09 \cellcolor[HTML]{DAE8FC} &97.40 \cellcolor[HTML]{DAE8FC} &100.00 \cellcolor[HTML]{DAE8FC} \\ 
\Xcline{2-3}{0.4pt}
\multicolumn{1}{c|}{}           & \multicolumn{1}{c|}{Common-}    & Micro &70.45 \cellcolor[HTML]{DAE8FC} &77.60 \cellcolor[HTML]{DAE8FC} &57.51 \cellcolor[HTML]{DAE8FC} &91.46 \cellcolor[HTML]{DAE8FC} &91.36 \cellcolor[HTML]{DAE8FC} \\ 
\Xcline{3-3}{0.4pt}
\multicolumn{1}{c|}{Test}       & \multicolumn{1}{c|}{sense}      & Macro &0.32 \cellcolor[HTML]{DAE8FC} &7.46 \cellcolor[HTML]{DAE8FC} &1.95 \cellcolor[HTML]{DAE8FC} &50.00 \cellcolor[HTML]{DAE8FC} &45.45 \cellcolor[HTML]{DAE8FC} \\ 
\Xcline{2-3}{0.4pt}
\multicolumn{1}{c|}{Set (308)}  & \multicolumn{1}{c|}{Hard}       & Micro &1.21 \cellcolor[HTML]{DAE8FC} &13.53 \cellcolor[HTML]{DAE8FC} &1.37 \cellcolor[HTML]{DAE8FC} &53.96 \cellcolor[HTML]{DAE8FC} &52.74 \cellcolor[HTML]{DAE8FC} \\ 
\Xcline{3-3}{0.4pt}
\multicolumn{1}{c|}{}           & \multicolumn{1}{c|}{Constraint} & Macro &0.32 \cellcolor[HTML]{DAE8FC} &9.74 \cellcolor[HTML]{DAE8FC} &1.30 \cellcolor[HTML]{DAE8FC} &45.12 \cellcolor[HTML]{DAE8FC} &29.22 \cellcolor[HTML]{DAE8FC} \\ 
\Xcline{2-3}{0.4pt}
\multicolumn{1}{c|}{}           & \multicolumn{2}{c|}{\colorr{Final Pass Rate}}    &0.00 \cellcolor[HTML]{DAE8FC} &2.92 \cellcolor[HTML]{DAE8FC} &0.65 \cellcolor[HTML]{DAE8FC} &\textbf{\colorb{42.68}} \cellcolor[HTML]{DAE8FC} &14.94 \cellcolor[HTML]{DAE8FC} \\ 
\Xhline{1.5pt}
\hline
\end{tabular}
}
\end{table*}

\begin{table*}[h]
\centering
\caption{The Average Pass Rates (\%) ``\textbf{Without} Unconventional Hint'' for Instances Across All Difficulty Level \zc{on TravelPlanner}. The highest final pass rates are highlighted in bold blue. \zc{For the last column ``LLaMA-3.1-8B + GPT-4 + PMC'', we employ LLaMA-3.1-8B as the planner and GPT-4 as the executor, where LLaMA-3.1-8B is prompted with one-shot example.}}
\label{tab:overall-performance-TravelPlanner-Without-commensense}
\setlength{\extrarowheight}{3pt}
\scalebox{0.6}{
\begin{tabular}{@{}ccc|cc|cc|c|c@{}}
\hline
\Xhline{1.5pt}
\multicolumn{3}{c|}{Without Unconventional Hint}                             & \zc{GPT-3.5} + ReAct          & \zc{GPT-4} + ReAct          & \zc{GPT-3.5} + \zc{PMC}            & \zc{GPT-4} + \zc{PMC}            & \zc{GPT-4} (SP)   & \zc{\xd{LLaMA-3.1-8B} + GPT-4 + PMC}       \\ 
\Xhline{1pt}
\multicolumn{1}{c|}{}           & \multicolumn{2}{c|}{Delivery Rate}      &100.00 \cellcolor[HTML]{EFEFEF} &98.33 \cellcolor[HTML]{EFEFEF} &88.33 \cellcolor[HTML]{EFEFEF} &100.00 \cellcolor[HTML]{EFEFEF} &100.00 \cellcolor[HTML]{EFEFEF} &70.00 \cellcolor[HTML]{EFEFEF} \\ 
\Xcline{2-3}{0.4pt}
\multicolumn{1}{c|}{}           & \multicolumn{1}{c|}{Common-}    & Micro &73.54 \cellcolor[HTML]{EFEFEF} &75.21 \cellcolor[HTML]{EFEFEF} &58.75 \cellcolor[HTML]{EFEFEF} &90.00 \cellcolor[HTML]{EFEFEF} &90.42 \cellcolor[HTML]{EFEFEF} &68.33 \cellcolor[HTML]{EFEFEF} \\ 
\Xcline{3-3}{0.4pt}
\multicolumn{1}{c|}{Validation} & \multicolumn{1}{c|}{sense}      & Macro &0.00 \cellcolor[HTML]{EFEFEF} &3.33 \cellcolor[HTML]{EFEFEF} &0.00 \cellcolor[HTML]{EFEFEF} &41.67 \cellcolor[HTML]{EFEFEF} &35.00 \cellcolor[HTML]{EFEFEF} &28.33 \cellcolor[HTML]{EFEFEF} \\ 
\Xcline{2-3}{0.4pt}
\multicolumn{1}{c|}{Set (60)}   & \multicolumn{1}{c|}{Hard}       & Micro &0.71 \cellcolor[HTML]{EFEFEF} &14.28 \cellcolor[HTML]{EFEFEF} &0.00 \cellcolor[HTML]{EFEFEF} &55.71 \cellcolor[HTML]{EFEFEF} &52.14 \cellcolor[HTML]{EFEFEF} &32.15 \cellcolor[HTML]{EFEFEF} \\ 
\Xcline{3-3}{0.4pt}
\multicolumn{1}{c|}{}           & \multicolumn{1}{c|}{Constraint} & Macro &1.67 \cellcolor[HTML]{EFEFEF} &13.33 \cellcolor[HTML]{EFEFEF} &0.00 \cellcolor[HTML]{EFEFEF} &48.33 \cellcolor[HTML]{EFEFEF} &25.00 \cellcolor[HTML]{EFEFEF} &25.00 \cellcolor[HTML]{EFEFEF} \\ 
\Xcline{2-3}{0.4pt}
\multicolumn{1}{c|}{}           & \multicolumn{2}{c|}{\colorr{Final Pass Rate}}    &0.00 \cellcolor[HTML]{EFEFEF} &0.00 \cellcolor[HTML]{EFEFEF} &0.00 \cellcolor[HTML]{EFEFEF} &\textbf{\colorb{31.67}} \cellcolor[HTML]{EFEFEF} &6.67 \cellcolor[HTML]{EFEFEF} &15.00 \cellcolor[HTML]{EFEFEF} \\ 
\Xhline{1.5pt}
\hline
\multicolumn{1}{c|}{}           & \multicolumn{2}{c|}{Delivery Rate}      &98.38 \cellcolor[HTML]{DAE8FC} &96.43 \cellcolor[HTML]{DAE8FC} &80.19 \cellcolor[HTML]{DAE8FC} &98.05 \cellcolor[HTML]{DAE8FC} &100.00 \cellcolor[HTML]{DAE8FC} &65.26 \cellcolor[HTML]{DAE8FC} \\ 
\Xcline{2-3}{0.4pt}
\multicolumn{1}{c|}{}           & \multicolumn{1}{c|}{Common-}    & Micro &69.60 \cellcolor[HTML]{DAE8FC} &70.74 \cellcolor[HTML]{DAE8FC} &51.54 \cellcolor[HTML]{DAE8FC} &85.96 \cellcolor[HTML]{DAE8FC} &88.07 \cellcolor[HTML]{DAE8FC} &57.35 \cellcolor[HTML]{DAE8FC} \\ 
\Xcline{3-3}{0.4pt}
\multicolumn{1}{c|}{Test}       & \multicolumn{1}{c|}{sense}      & Macro &0.00 \cellcolor[HTML]{DAE8FC} &2.92 \cellcolor[HTML]{DAE8FC} &0.65 \cellcolor[HTML]{DAE8FC} &29.55 \cellcolor[HTML]{DAE8FC} &32.79 \cellcolor[HTML]{DAE8FC} &21.43 \cellcolor[HTML]{DAE8FC} \\ 
\Xcline{2-3}{0.4pt}
\multicolumn{1}{c|}{Set (308)}  & \multicolumn{1}{c|}{Hard}       & Micro &1.37 \cellcolor[HTML]{DAE8FC} &14.44 \cellcolor[HTML]{DAE8FC} &0.91 \cellcolor[HTML]{DAE8FC} &50.91 \cellcolor[HTML]{DAE8FC} &50.91 \cellcolor[HTML]{DAE8FC} &23.71 \cellcolor[HTML]{DAE8FC} \\ 
\Xcline{3-3}{0.4pt}
\multicolumn{1}{c|}{}           & \multicolumn{1}{c|}{Constraint} & Macro &0.65 \cellcolor[HTML]{DAE8FC} &9.09 \cellcolor[HTML]{DAE8FC} &0.32 \cellcolor[HTML]{DAE8FC} &46.10 \cellcolor[HTML]{DAE8FC} &30.84 \cellcolor[HTML]{DAE8FC} &19.16 \cellcolor[HTML]{DAE8FC} \\ 
\Xcline{2-3}{0.4pt}
\multicolumn{1}{c|}{}           & \multicolumn{2}{c|}{\colorr{Final Pass Rate}}    &0.00 \cellcolor[HTML]{DAE8FC} &0.65 \cellcolor[HTML]{DAE8FC} &0.32 \cellcolor[HTML]{DAE8FC} &\textbf{\colorb{22.40}} \cellcolor[HTML]{DAE8FC} &12.66 \cellcolor[HTML]{DAE8FC} &12.66 \cellcolor[HTML]{DAE8FC} \\ 
\Xhline{1.5pt}
\hline
\end{tabular}
}
\end{table*}



\begin{table*}[h]
\centering
\caption{The Pass Rates Constraints for the ``With Unconventional Hint'' scenario \zc{for TravelPlanner}. 
The ``-'' marks indicate the corresponding constraints are not applicable.
The highest final pass rates are highlighted in bold blue for each difficulty level.}
\label{tab:detailed-constraint-pass-rate}
\setlength{\extrarowheight}{3pt}
\scalebox{0.6}{
\begin{tabular}{@{}cccccccccc@{}}
\Xhline{1.5pt}
\multicolumn{1}{c|}{}                                  & \multicolumn{3}{c|}{\zc{GPT4} + ReAct + CoT}                                                                      & \multicolumn{3}{c|}{Sole-planning}                                                                  & \multicolumn{3}{c}{\zc{PMC}}                            \\ 
\Xcline{2-10}{0.4pt}
\multicolumn{1}{c|}{\multirow{-2}{*}{Constraint Type}} & Easy                     & Medium                   & \multicolumn{1}{c|}{Hard}                     & Easy                     & Medium                   & \multicolumn{1}{c|}{Hard}                     & Easy                     & Medium                   & Hard                     \\ 
\Xhline{1pt}
\multicolumn{10}{c}{\cellcolor[HTML]{EFEFEF}\textbf{Commonsense Constraint}}                                                                                                                                                                                                                                                                                 \\ 
\Xhline{1pt}
\multicolumn{1}{c|}{Within Sandbox}                    &32.79 \cellcolor[HTML]{DAE8FC} &23.08 \cellcolor[HTML]{DAE8FC} & \multicolumn{1}{c|}{\cellcolor[HTML]{DAE8FC}21.95} &90.16 \cellcolor[HTML]{E7E8FC} &88.46   \cellcolor[HTML]{E7E8FC} & \multicolumn{1}{c|}{\cellcolor[HTML]{E7E8FC}82.93} &70.49 \cellcolor[HTML]{E1F5E0} &75.96 \cellcolor[HTML]{E1F5E0} &70.73 \cellcolor[HTML]{E1F5E0} \\
\multicolumn{1}{c|}{Complete Information}              &81.97 \cellcolor[HTML]{DAE8FC} &86.54 \cellcolor[HTML]{DAE8FC} & \multicolumn{1}{c|}{\cellcolor[HTML]{DAE8FC}86.59} &100.00 \cellcolor[HTML]{E7E8FC} &100.00 \cellcolor[HTML]{E7E8FC} & \multicolumn{1}{c|}{\cellcolor[HTML]{E7E8FC}100} &79.51 \cellcolor[HTML]{E1F5E0} &78.85 \cellcolor[HTML]{E1F5E0} &79.27 \cellcolor[HTML]{E1F5E0} \\
\multicolumn{1}{c|}{Within Current City}               &95.08 \cellcolor[HTML]{DAE8FC} &95.19 \cellcolor[HTML]{DAE8FC} & \multicolumn{1}{c|}{\cellcolor[HTML]{DAE8FC}90.24} &97.54 \cellcolor[HTML]{E7E8FC} &95.19 \cellcolor[HTML]{E7E8FC} & \multicolumn{1}{c|}{\cellcolor[HTML]{E7E8FC}98.78} &97.54 \cellcolor[HTML]{E1F5E0} &95.19 \cellcolor[HTML]{E1F5E0} &97.56 \cellcolor[HTML]{E1F5E0} \\
\multicolumn{1}{c|}{Reasonable City Route}             &88.52 \cellcolor[HTML]{DAE8FC} &88.46 \cellcolor[HTML]{DAE8FC} & \multicolumn{1}{c|}{\cellcolor[HTML]{DAE8FC}87.80} &100.00 \cellcolor[HTML]{E7E8FC} &99.04 \cellcolor[HTML]{E7E8FC} & \multicolumn{1}{c|}{\cellcolor[HTML]{E7E8FC}100} &94.26 \cellcolor[HTML]{E1F5E0} &93.27 \cellcolor[HTML]{E1F5E0} &97.56 \cellcolor[HTML]{E1F5E0} \\
\multicolumn{1}{c|}{Diverse Restaurants}               &81.15 \cellcolor[HTML]{DAE8FC} &76.92 \cellcolor[HTML]{DAE8FC} & \multicolumn{1}{c|}{\cellcolor[HTML]{DAE8FC}71.95} &98.36 \cellcolor[HTML]{E7E8FC} &86.54 \cellcolor[HTML]{E7E8FC} & \multicolumn{1}{c|}{\cellcolor[HTML]{E7E8FC}95.12} &88.52 \cellcolor[HTML]{E1F5E0} &83.65 \cellcolor[HTML]{E1F5E0} &97.56 \cellcolor[HTML]{E1F5E0} \\
\multicolumn{1}{c|}{Diverse Attractions}               &99.18 \cellcolor[HTML]{DAE8FC} &98.08 \cellcolor[HTML]{DAE8FC} & \multicolumn{1}{c|}{\cellcolor[HTML]{DAE8FC}96.34} &100.00 \cellcolor[HTML]{E7E8FC} &100.00 \cellcolor[HTML]{E7E8FC} & \multicolumn{1}{c|}{\cellcolor[HTML]{E7E8FC}100} &97.54 \cellcolor[HTML]{E1F5E0} &95.19 \cellcolor[HTML]{E1F5E0} &100.00 \cellcolor[HTML]{E1F5E0} \\
\multicolumn{1}{c|}{Non-conf. Transportation}         &93.44 \cellcolor[HTML]{DAE8FC} &96.15 \cellcolor[HTML]{DAE8FC} & \multicolumn{1}{c|}{\cellcolor[HTML]{DAE8FC}95.12} &95.08 \cellcolor[HTML]{E7E8FC} &95.19 \cellcolor[HTML]{E7E8FC} & \multicolumn{1}{c|}{\cellcolor[HTML]{E7E8FC}97.56} &93.44 \cellcolor[HTML]{E1F5E0} &92.31 \cellcolor[HTML]{E1F5E0} &100.00 \cellcolor[HTML]{E1F5E0} \\
\multicolumn{1}{c|}{Miminum Nights Stay}               &57.38 \cellcolor[HTML]{DAE8FC} &67.31 \cellcolor[HTML]{DAE8FC} & \multicolumn{1}{c|}{\cellcolor[HTML]{DAE8FC}43.90} &62.30 \cellcolor[HTML]{E7E8FC} &53.85 \cellcolor[HTML]{E7E8FC} & \multicolumn{1}{c|}{\cellcolor[HTML]{E7E8FC}53.66} &95.90 \cellcolor[HTML]{E1F5E0} &93.27 \cellcolor[HTML]{E1F5E0} &89.02 \cellcolor[HTML]{E1F5E0} \\

\Xhline{1pt}
\multicolumn{10}{c}{\cellcolor[HTML]{EFEFEF}\textbf{Hard Constraint}}                                                                                                                                                                                                                                                                                        \\ 
\Xhline{1pt}
\multicolumn{1}{c|}{Budget}                            &13.93 \cellcolor[HTML]{DAE8FC} &10.58 \cellcolor[HTML]{DAE8FC} & \multicolumn{1}{c|}{\cellcolor[HTML]{DAE8FC}9.76} &50.00 \cellcolor[HTML]{E7E8FC} &34.62 \cellcolor[HTML]{E7E8FC} & \multicolumn{1}{c|}{\cellcolor[HTML]{E7E8FC}14.63} &46.72 \cellcolor[HTML]{E1F5E0} &42.31 \cellcolor[HTML]{E1F5E0} &46.34 \cellcolor[HTML]{E1F5E0} \\
\multicolumn{1}{c|}{Room Rule}                         &- \cellcolor[HTML]{DAE8FC} &13.16 \cellcolor[HTML]{DAE8FC} & \multicolumn{1}{c|}{\cellcolor[HTML]{DAE8FC}17.81} &- \cellcolor[HTML]{E7E8FC} &47.37 \cellcolor[HTML]{E7E8FC} & \multicolumn{1}{c|}{\cellcolor[HTML]{E7E8FC}65.75} &- \cellcolor[HTML]{E1F5E0} &57.89 \cellcolor[HTML]{E1F5E0} &57.53 \cellcolor[HTML]{E1F5E0} \\
\multicolumn{1}{c|}{Cuisine}                           &- \cellcolor[HTML]{DAE8FC} &11.43 \cellcolor[HTML]{DAE8FC} & \multicolumn{1}{c|}{\cellcolor[HTML]{DAE8FC}13.51} &- \cellcolor[HTML]{E7E8FC} &65.71 \cellcolor[HTML]{E7E8FC} & \multicolumn{1}{c|}{\cellcolor[HTML]{E7E8FC}45.95} &- \cellcolor[HTML]{E1F5E0} &62.86 \cellcolor[HTML]{E1F5E0} &51.35 \cellcolor[HTML]{E1F5E0} \\
\multicolumn{1}{c|}{Room Type}                         &- \cellcolor[HTML]{DAE8FC} &19.35 \cellcolor[HTML]{DAE8FC} & \multicolumn{1}{c|}{\cellcolor[HTML]{DAE8FC}14.29} &- \cellcolor[HTML]{E7E8FC} &74.19 \cellcolor[HTML]{E7E8FC} & \multicolumn{1}{c|}{\cellcolor[HTML]{E7E8FC}77.78} &- \cellcolor[HTML]{E1F5E0} &54.84 \cellcolor[HTML]{E1F5E0} &53.97 \cellcolor[HTML]{E1F5E0} \\
\multicolumn{1}{c|}{Transportation}                    &- \cellcolor[HTML]{DAE8FC} &- \cellcolor[HTML]{DAE8FC} & \multicolumn{1}{c|}{\cellcolor[HTML]{DAE8FC}15.07} &- \cellcolor[HTML]{E7E8FC} &- \cellcolor[HTML]{E7E8FC} & \multicolumn{1}{c|}{\cellcolor[HTML]{E7E8FC}82.19} &- \cellcolor[HTML]{E1F5E0} &- \cellcolor[HTML]{E1F5E0} &60.27 \cellcolor[HTML]{E1F5E0} \\
\Xhline{1pt}
\multicolumn{10}{c}{\cellcolor[HTML]{EFEFEF}\textbf{Final}}                                                                                                                                                                                                                                                                                                  \\ 
\Xhline{1pt}
\multicolumn{1}{c|}{\colorr{Final Pass Rate}}                   &4.10 \cellcolor[HTML]{DAE8FC} &4.10 \cellcolor[HTML]{DAE8FC} & \multicolumn{1}{c|}{\cellcolor[HTML]{DAE8FC}2.44} &30.33 \cellcolor[HTML]{E7E8FC} &5.77 \cellcolor[HTML]{E7E8FC} & \multicolumn{1}{c|}{\cellcolor[HTML]{E7E8FC}3.66} &\textbf{\colorb{43.44}} \cellcolor[HTML]{E1F5E0} &\textbf{\colorb{34.62}} \cellcolor[HTML]{E1F5E0} &\textbf{\colorb{42.68}} \cellcolor[HTML]{E1F5E0} \\ 
\Xhline{1.5pt}
\end{tabular}
}
\end{table*}


\subsection{Hierarchical Task Planning and Execution}\label{hierarchical}

\paragraph{Task-Level Planning.}
The manager agent will analyze the given complex task $\mathcal{T}$ and decompose it into a series of inter-dependent \zc{sub-task}s $\{T_1, T_2, \dots\}$. \zc{PMC} represents them via a directed graph $\mathcal{G}=(\mathcal{V}, \mathcal{E})$, termed \textit{\zc{sub-task} graph}. In $\mathcal{G}$, each node $V_i \in \mathcal{V}$ corresponds to a \zc{sub-task} $T_i$ and each edge $E_{ij} \in \mathcal{E}$ delineates the dependencies between \zc{sub-task}s $T_i$ and $T_j$, where $i,j \in S(K)$. The architecture of the \zc{sub-task} graph is illustrated in Figure~\ref{fig:meta-task-graph}. Executors adhere to the graph's topology, \ie its edge orientation, to ensure all prerequisites of a \zc{sub-task} are met before its initiation,
thereby enhancing the efficacy of the overall task execution. 
Furthermore, the \zc{sub-task} graph serves as a tool for visualizing the task decomposition and an interactive interface to enhance the interpretability of \zc{PMC} systems. It offers a mechanism for ongoing monitoring and potential human intervention, making it essential for \zc{PMC}.

\paragraph{Step-Level Planning and Execution.}
After task-level decomposition, each \zc{sub-task} is manageable to be further decomposed into a sequence of executable actions, \ie function calls. The complexity of each \zc{sub-task} is significantly reduced as it is now a specific, well-defined task with clear requirements and local constraints, making the off-the-shelf planning method directly applicable, \eg ReAct~\cite{yao2023react}. Specifically, before the commencement of \zc{sub-task} $T_i$, the supervisor agent will rewrite $T_i$ by referencing the outcomes of all neighboring \zc{sub-task}s $\mathcal{N}(T_i)$. Then, the local constraints $C_l$ for \zc{sub-task} $T_i$ identified by the manager will be given as the auxiliary information together with the refined $T_i$ to the executor agent $A_i$, which will utilize the planning method, \eg ReAct, to accomplish $T_i$ by decoding $T_i$ into a sequence of actions. The whole process is illustrated in Figure~\ref{fig:step-level-planning-and-execution}.

\section{Experiments}

To assess \zc{PMC}, we move beyond the existing planning methods that largely focus on simplistic tasks~\cite{singh2023progprompt} or puzzles~\cite{ahn2024large} irrelevant to practical applications. Instead, we evaluate \zc{PMC} through its application to real-world \zc{constraints intensive} scenarios. Specifically, we examine its efficacy in the domains of itinerary planning~\cite{xie2024travelplanner} and daily tools using~\cite{li2023api}. We \zc{optimized \zc{PMC}'s prompt} for each benchmark exclusively on the validation set and apply the \zc{prompt} directly to the test sets. For~\cite{li2023api}, the training dataset was employed as a proxy for the validation set owing to the lack of a dedicated validation set. For all benchmarks, each instance is executed only once without sampling, though multiple trials could potentially enhance performance. We will make the code publicly available upon acceptance.

\subsection{Experiment Setup}\label{ssec:exp_setup}

\subsubsection{Benchmarks}
\textbf{TravelPlanner}~\cite{xie2024travelplanner}. In \texttt{TravelPlanner}, users specify their origin, destination, and individual requirements. The benchmark assesses the ability of language agents to (1) efficiently gather necessary information using appropriate tools and (2) create practical, personalized travel plans for users. The plan is assessed using four main metrics: (1) delivery rate (a plan has to be delivered within \zc{\textbf{30 steps (including planning and execution)}}), (2) commonsense constraint pass rate, (3) hard constraint pass rate, and (4) final pass rate (the rate for meeting all commonsense and hard constraints), which is the most important metric for evaluation. For (2) and (3), we define the "micro" pass rate as the ratio of passed constraints to total constraints and the "macro" pass rate as the ratio of plans passing all constraints to total plans.

The travel duration can be 3, 5, or 7 days. Due to budget constraints, we demonstrate that a 3-day dataset sufficiently justifies the effectiveness of \zc{PMC}. The queries are categorized as \texttt{easy}, \texttt{medium} or \texttt{hard}. 

However, we found that the benchmark includes odd rules as part of its evaluation. For instance, choosing the same restaurant multiple times throughout a trip breaches the \texttt{Diverse Restaurants} constraint, and selecting an airport as a meal location breaches the \texttt{Within Sandbox} constraint. Yet, under normal circumstances, it's reasonable for a tourist to return to a favoured restaurant or dine at airport restaurants during their trip. To ensure that the agent recognizes these rules as part of commonsense knowledge, we provide specific guidance to the planning agents: the \texttt{Deliverer Agent} in \zc{PMC} and the \texttt{Planner} in React and Sole-Planning. \zc{We term this setting as ``\textbf{with unconventional hint}'' (or with ``hint'' for short).} To maintain the integrity of the experiment and stay true to the objectives of the original \texttt{TravelPlanner} \zc{paper}, we conduct a separate experiment that excludes this external knowledge. This experiment still incorporates the less conventional rules used in both \texttt{Diverse Restaurants} and \texttt{Within Sandbox} settings. \zc{We term this setting as ``\textbf{without unconventional hint}'' (or without ``hint'' for short).}

As our method consists of tool-use and planning (two-stage), we compare our method with the two-stage baseline, \texttt{ReAct} from \cite{xie2024travelplanner} using \zc{GPT-3.5}-Turbo and \zc{GPT-4}-Turbo as language models. We also further compare our method to the best sole-planning baseline, Direct \zc{GPT4}-Turbo, which has provided necessary information to the agent and only require agent to output the travel plan.

\textbf{API-Bank}~\cite{li2023api}. \texttt{API-Bank} is a benchmark designed to evaluate the tool-use capabilities of large language models, focusing on APIs that are commonly used in everyday life, such as email. The benchmark includes three levels of difficulty, with Level 3 being the most challenging. We chose Level 3 for our experiment because it best assesses the planning abilities of the agent.

The benchmark assesses agents based on Accuracy and ``\textbf{ROUGE}'' (ROUGE-L) scores. The Accuracy metric gauges the correctness of API calls based on user queries, calculated as the proportion of correct API calls to total predictions. We modified this metric for a more consistent and fair assessment by defining Correctness as the ratio of unique correct API calls to total predictions. This adjustment addresses the tendency of some language models, like \zc{GPT-3.5} and \zc{GPT-4}, to make repetitive correct API calls. The ROUGE-L score evaluates the responses generated from these API calls. Our experiments indicate that using this refined Accuracy metric results in lower baseline scores.

In addition to Correctness, we introduce the ``\textbf{Completeness}'' to better assess task execution. Correctness alone may not fully capture an agent's performance, as minimal API calls could artificially inflate scores. Completeness measures the ratio of unique, correct API calls to the total required API calls for the task, addressing the limitations of Correctness and ensuring a more accurate evaluation of the agent's effectiveness. We also introduce another metric named ``\textbf{Tool Repeats}'',  which measures how often the model correctly calls an API after its initial use. A lower number of repeats indicates fewer unnecessary inferences, signifying a more efficient solution.


\begin{table*}[t]
\centering
\caption{The Performance on API-Bank. The highest performance for each criteria is highlighted in bold blue. \zc{For the last column ``LLaMA-3.1-8B + GPT-4'', we employ LLaMA-3.1-8B as the planner and GPT-4 as the executor, where LLaMA-3.1-8B is prompted with one-shot example.}}
\label{performance-on-API-Bank}
\setlength{\extrarowheight}{3pt}
\scalebox{0.6}{
\begin{tabular}{@{}cc|ccccc@{}}
\Xhline{1.5pt}
\multicolumn{2}{c|}{Model}                                   & \zc{GPT-4}-1106-Preview       & \zc{GPT-4}-0613               & $\texttt{GPT}$-3.5-Turbo-0125       & $\texttt{GPT}$-3.5-Turbo-0613 & \zc{LLaMA-3.1-8B} + GPT-4   \\ 
\Xhline{1pt}
\multicolumn{1}{c|}{}                               & CoT    &71.48 \cellcolor[HTML]{EFEFEF} &57.65 \cellcolor[HTML]{EFEFEF} &35.38 \cellcolor[HTML]{EFEFEF} &34.74 \cellcolor[HTML]{EFEFEF} & - \cellcolor[HTML]{EFEFEF} \\ 
\Xcline{2-2}{0.4pt}
\multicolumn{1}{c|}{}                               & No CoT &41.58 \cellcolor[HTML]{EFEFEF} &35.11 \cellcolor[HTML]{EFEFEF} &67.06 \cellcolor[HTML]{EFEFEF} &41.80 \cellcolor[HTML]{EFEFEF} & - \cellcolor[HTML]{EFEFEF}  \\ 
\Xcline{2-2}{0.4pt}
\multicolumn{1}{c|}{\multirow{-3}{*}{Correctness \% ($\uparrow$)}}  & \zc{PMC}    &82.63 \cellcolor[HTML]{EFEFEF} &\textbf{\colorb{85.12}} \cellcolor[HTML]{EFEFEF} &74.13 \cellcolor[HTML]{EFEFEF} &67.81 \cellcolor[HTML]{EFEFEF} &67.58 \cellcolor[HTML]{EFEFEF} \\ 
\Xhline{1pt}
\multicolumn{1}{c|}{}                               & CoT    &47.76 \cellcolor[HTML]{DAE8FC} &46.12 \cellcolor[HTML]{DAE8FC} &28.16 \cellcolor[HTML]{DAE8FC} &30.20 \cellcolor[HTML]{DAE8FC} &- \cellcolor[HTML]{DAE8FC} \\ 
\Xcline{2-2}{0.4pt}
\multicolumn{1}{c|}{}                               & No CoT &34.29 \cellcolor[HTML]{DAE8FC} &32.24 \cellcolor[HTML]{DAE8FC} &23.27 \cellcolor[HTML]{DAE8FC} &20.82 \cellcolor[HTML]{DAE8FC} &- \cellcolor[HTML]{DAE8FC} \\ 
\Xcline{2-2}{0.4pt}
\multicolumn{1}{c|}{\multirow{-3}{*}{Completeness \% ($\uparrow$)}} & \zc{PMC}    &\textbf{\colorb{64.08}} \cellcolor[HTML]{DAE8FC} &58.37 \cellcolor[HTML]{DAE8FC} &43.27 \cellcolor[HTML]{DAE8FC} &40.40 \cellcolor[HTML]{DAE8FC} &70.61 \cellcolor[HTML]{DAE8FC} \\ 
\Xhline{1pt}
\multicolumn{1}{c|}{}                               & CoT    &0.2641 \cellcolor[HTML]{E7E8FC} &0.2846 \cellcolor[HTML]{E7E8FC} &0.3085 \cellcolor[HTML]{E7E8FC} &0.2656 \cellcolor[HTML]{E7E8FC} &- \cellcolor[HTML]{E7E8FC} \\ 
\Xcline{2-2}{0.4pt}
\multicolumn{1}{c|}{}                               & No CoT &0.2507 \cellcolor[HTML]{E7E8FC} &0.2644 \cellcolor[HTML]{E7E8FC} &0.2346 \cellcolor[HTML]{E7E8FC} &0.2016 \cellcolor[HTML]{E7E8FC} &- \cellcolor[HTML]{E7E8FC} \\ 
\Xcline{2-2}{0.4pt}
\multicolumn{1}{c|}{\multirow{-3}{*}{ROUGE ($\uparrow$)}}        & \zc{PMC}    &\textbf{\colorb{0.4053}} \cellcolor[HTML]{E7E8FC} &0.3839 \cellcolor[HTML]{E7E8FC} &0.3894 \cellcolor[HTML]{E7E8FC} &0.3754 \cellcolor[HTML]{E7E8FC} &0.4171 \cellcolor[HTML]{E7E8FC} \\ 
\Xhline{1pt}
\multicolumn{1}{c|}{}                               & CoT    &90 \cellcolor[HTML]{E1F5E0} &49 \cellcolor[HTML]{E1F5E0} &50 \cellcolor[HTML]{E1F5E0} &32 \cellcolor[HTML]{E1F5E0} &- \cellcolor[HTML]{E1F5E0}  \\ 
\Xcline{2-2}{0.4pt}
\multicolumn{1}{c|}{}                               & No CoT &38 \cellcolor[HTML]{E1F5E0} &15 \cellcolor[HTML]{E1F5E0} &155 \cellcolor[HTML]{E1F5E0} &118 \cellcolor[HTML]{E1F5E0}  &- \cellcolor[HTML]{E1F5E0}\\ 
\Xcline{2-2}{0.4pt}
\multicolumn{1}{c|}{\multirow{-3}{*}{Tool Repeats ($\downarrow$)}}      & \zc{PMC}    &16 \cellcolor[HTML]{E1F5E0} &23 \cellcolor[HTML]{E1F5E0} &\textbf{\colorb{7}} \cellcolor[HTML]{E1F5E0} &65 \cellcolor[HTML]{E1F5E0} &64 \cellcolor[HTML]{E1F5E0} \\ 
\Xhline{1.5pt}
\end{tabular}
}
\end{table*}


\subsection{Result Analysis}

\subsubsection{Result Analysis for \texttt{TravelPlanner}}

From Tables~\ref{tab:overall-performance-TravelPlanner-with-commensense} and ~\ref{tab:overall-performance-TravelPlanner-Without-commensense}, it is evident that \zc{PMC} significantly outperforms all baseline methods irrespective of the presence of unconventional hints. Notably, when hints are included, \zc{GPT4} enhanced by \zc{PMC} achieves a superior average final pass rate of 42.68\% across all difficulty levels, compared to a meagre 2.92\% by baselines. This data underscores the potential of integrating large language models (LLMs) with multi-agent systems, marking it as a promising area for future research in LLM-based agent systems.

In the absence of hints, the setting replicates that described in~\cite{xie2024travelplanner}, where the highest final pass rate for baseline models stands at 0.56\%, consistent with the original study's findings. In this scenario, \zc{PMC} significantly improves with an average final pass rate of 22.4\%, surpassing the best-reported baseline result in~\cite{xie2024travelplanner}.

Notably, the \zc{sub-task} Planner (\zc{PMC}) significantly outperforms the Standard Planner (SP) in settings that employ hints and those that do not. The SP operates purely as a decision-making framework in which all elements necessary for completing the itinerary, such as multiple choices for hotels, flights, and restaurants, are pre-supplied; thus, the SP agent merely selects the most suitable options from these pre-defined sets to construct the final itinerary. This renders SP a relatively simpler task compared to \zc{PMC} and other benchmarks, which necessitate the searching and gathering of necessary elements prior to decision-making. Nonetheless, \zc{PMC} achieves a superior final pass rate, a finding which may appear counter-intuitive yet can be elucidated as follows: \zc{PMC}'s exceptional performance is attributable to its effective deployment of a divide-and-conquer strategy in managing constraints. By resolving numerous local constraints during the execution of \zc{sub-task}s, \zc{PMC} considerably reduces the complexity that the agent encounters in formulating the ultimate itinerary plan. Table~\ref{tab:detailed-constraint-pass-rate} presents the detailed pass rates for individual constraints, indicating that \zc{PMC} significantly outperforms \zc{GPT4}+ReAct+CoT in terms of pass rates across all constraints. However, 
\zc{GPT-3.5}
is less effective than \zc{GPT4} when equipped with \zc{PMC}, possibly due to less model performance. We have provided illustrative results for each difficulty level in Appendix~\ref{appendix:result-demo}. We also present the detailed results for each difficulty level (easy, medium, hard) in Appendix~\ref{appendix:more-results}.

\subsubsection{Result Analysis for \texttt{API-Bank}}

Firstly, \zc{PMC} significantly enhances the performance of both \zc{GPT4} and 
\zc{GPT-3.5}
across all critical evaluation metrics. Compared to existing baselines, \zc{PMC} consistently demonstrates superior performance. Notably, the best performance reported in the original paper~\cite{li2023api} achieved a 70\% success rate, which our reimplementation slightly exceeds at 71.48\%. Thus, \zc{PMC} stands out by surpassing the top method referenced in ~\cite{li2023api} by a substantial margin of at least 14\%. Furthermore, \zc{PMC} excels in other key areas such as task completeness, achieving an impressive 64.08\%, and exhibits significantly fewer redundant tool interactions, with a count of just seven. This robust performance underscores \zc{PMC}'s potential in redefining the capabilities of advanced language models. We have provided illustrative results for each difficulty level in Appendix~\ref{appendix:result-demo}.

\subsubsection{\zc{Results for \xd{LLaMA-3.1-8B}}}
\zc{In addition to utilising only GPT models on PMC, we also test the performance of PMC with open-source LLM with significantly lesser parameters, i.e., LLaMA-3.1-8B. Specifically, we use LLaMA-3.1-8B as the planner agent, with prompt consists of one-shot example and without unconventional hint setting. The results for TravelPlanner and API-Bank are given in the last column in Table~\ref{tab:overall-performance-TravelPlanner-Without-commensense} and Table~\ref{performance-on-API-Bank}.}

\zc{From the results we demonstrate that even with small open-source LLM as the planner in PMC, its performance still surpasses GPT-4, with ReAct or CoT by a large margin. These results demonstrate that PMC is still effective for small LLMs.}

\section{Conclusion}
This paper presents \zc{Planning with Multi-Constraints (PMC)}, an innovative zero-shot methodology for collaborative LLM-based multi-agent systems. \zc{PMC} simplifies complex task planning by breaking it down into hierarchical \zc{sub-task}s, each mapped to executable actions. \zc{PMC} was evaluated on two benchmarks, TravelPlanner and API-Bank. It achieved an average success rate of about 42\% on TravelPlanner, a significant improvement from the initial 0.6\%, and outperformed \zc{GPT-4} with ReAct by 14\% on API-Bank. However, the current design still requires human input from executor agents. Enhancing \zc{PMC} by enabling the manager agent to autonomously design prompts for executor agents could optimize executor creation, accelerating \zc{PMC}'s practical application. Future research should focus on developing more autonomous agents through advanced prompt optimization, as suggested in recent literature~\cite{yang2023large}. This approach promises to refine \zc{PMC}'s functionality and expand its applicability without human intervention, leading to more intelligent and self-sufficient multi-agent systems.


\bibliography{custom}

\appendix

\section{Limitations}\label{appendix:limitations}
Despite the remarkable performance, the current architecture of executor agents in \zc{PMC} still necessitates human input. There is an opportunity for significant enhancements within \zc{PMC} by enabling the manager agent to autonomously generate the prompts for the executor agents, thereby optimizing the process of executor creation. Such advancements could notably accelerate the practical applications and enhance the efficacy of \zc{PMC}.

\section{Social Impacts}\label{appendix:broader_impacts}
This research explores the implementation of LLM-based agents to aid humans in solving complex tasks. While this automation promises increased productivity and focus on high-level tasks, it raises concerns about transparency and interpretability. The uncertainty of LLMs can obscure decision-making processes, potentially reducing trust in sensitive applications where understanding AI's reasoning is crucial.  Additionally, by automating communication, there is a risk of diminishing important human interactions and nuances essential for quality exchanges and relationships in professional environments. It is imperative that the deployment of these agents incorporates ethical considerations and mechanisms for explainability to mitigate these risks, ensuring they contribute positively and responsibly to societal and technological advancements.

\section{The Overview of Prompt for Each Agent}

\section{\zc{Proof of Proposition~\ref{proposition: 1}}}\label{proof: proposition 1}
\begin{proposition}
    \textit{A \zc{sub-task} $T_i$ is accomplishable while adhering to local constraints if and only if all the \zc{sub-task}s within its direct neighborhood $\mathcal{N}(T_i)$ are accomplished with their respective local constraints maintained.}
\end{proposition}

The proof is straightforward. If all tasks $T_j \in \mathcal{N}(T_i)$ are accomplished, then all the prerequisite requirements for $T_i$ are satisfied (since $T_i$ only depends on $\mathcal{N}(T_i)$), enabling the completion of $T_i$. Conversely, assume $T_i$ is accomplishable even if one of its prerequisite tasks $T_j$ (a direct neighbor or connected via a path $Path_{ji}$ to $T_i$) fails, this failure would propagate recursively to $T_i$, inevitably leading to $N_i$'s failure, contradicting with the assumption that $T_i$ is accomplishable. The supervisor prompt design is delineated in Figure~\ref{fig:agent-design-supervisor}.

\section{More Results}\label{appendix:more-results}

\subsection{Result for TravelPlanner with Hint for Each Difficulty Level.}

\zc{Please refer to Table~\ref{performance-on-easy-instances-TravelPlanner-with-commensense}, Table~\ref{performance-on-medium-instances-TravelPlanner-with-commensense}, and Table~\ref{performance-on-hard-instances-TravelPlanner-with-commensense} for detailed results for each difficulty level on TravelPlanner.}


\begin{table*}[h]
\centering
\caption{The Pass Rates (\%) on \textbf{Easy} Instances. The highest final pass rates are highlighted in bold blue.}
\label{performance-on-easy-instances-TravelPlanner-with-commensense}
\setlength{\extrarowheight}{3pt}
\scalebox{0.7}{
\begin{tabular}{@{}ccc|cc|cc|c@{}}
\hline
\Xhline{1.5pt}
\multicolumn{3}{c|}{With Unconventional Hint}                             & \zc{GPT-3.5} + ReAct          & \zc{GPT-4} + ReAct          & \zc{GPT-3.5} + \zc{PMC}            & \zc{GPT-4} + \zc{PMC}            & \zc{GPT-4} (SP)             \\ 
\Xhline{1pt}
\multicolumn{1}{c|}{}           & \multicolumn{2}{c|}{Delivery Rate}      &100.00 \cellcolor[HTML]{EFEFEF} &95.00 \cellcolor[HTML]{EFEFEF} &90.00 \cellcolor[HTML]{EFEFEF} &100.00 \cellcolor[HTML]{EFEFEF} &100.00 \cellcolor[HTML]{EFEFEF} \\ 
\Xcline{2-3}{0.4pt}
\multicolumn{1}{c|}{}           & \multicolumn{1}{c|}{Common-}    & Micro &76.88 \cellcolor[HTML]{EFEFEF} &75.00 \cellcolor[HTML]{EFEFEF} &61.88 \cellcolor[HTML]{EFEFEF} &95.63 \cellcolor[HTML]{EFEFEF} &95.00 \cellcolor[HTML]{EFEFEF} \\ 
\Xcline{3-3}{0.4pt}
\multicolumn{1}{c|}{Validation} & \multicolumn{1}{c|}{sense}      & Macro &0.00 \cellcolor[HTML]{EFEFEF} &5.00 \cellcolor[HTML]{EFEFEF} &0.00 \cellcolor[HTML]{EFEFEF} &70.00 \cellcolor[HTML]{EFEFEF} &65.00 \cellcolor[HTML]{EFEFEF} \\ 
\Xcline{2-3}{0.4pt}
\multicolumn{1}{c|}{Set (20)}   & \multicolumn{1}{c|}{Hard}       & Micro &0.00 \cellcolor[HTML]{EFEFEF} &15.00 \cellcolor[HTML]{EFEFEF} &5.00 \cellcolor[HTML]{EFEFEF} &55.00 \cellcolor[HTML]{EFEFEF} &60.00 \cellcolor[HTML]{EFEFEF} \\ 
\Xcline{3-3}{0.4pt}
\multicolumn{1}{c|}{}           & \multicolumn{1}{c|}{Constraint} & Macro &0.00 \cellcolor[HTML]{EFEFEF} &15.00 \cellcolor[HTML]{EFEFEF} &5.00 \cellcolor[HTML]{EFEFEF} &55.00 \cellcolor[HTML]{EFEFEF} &60.00 \cellcolor[HTML]{EFEFEF} \\ 
\Xcline{2-3}{0.4pt}
\multicolumn{1}{c|}{}           & \multicolumn{2}{c|}{\colorr{Final Pass Rate}}    &0.00 \cellcolor[HTML]{EFEFEF} &5.00 \cellcolor[HTML]{EFEFEF} &0.00 \cellcolor[HTML]{EFEFEF} &\textbf{\colorb{55.00}} \cellcolor[HTML]{EFEFEF} &35.00 \cellcolor[HTML]{EFEFEF} \\ 
\Xhline{1.5pt}
\hline
\multicolumn{1}{c|}{}           & \multicolumn{2}{c|}{Delivery Rate}      &90.16 \cellcolor[HTML]{DAE8FC} &99.18 \cellcolor[HTML]{DAE8FC} &90.16 \cellcolor[HTML]{DAE8FC} &97.54 \cellcolor[HTML]{DAE8FC} &100.00 \cellcolor[HTML]{DAE8FC} \\ 
\Xcline{2-3}{0.4pt}
\multicolumn{1}{c|}{}           & \multicolumn{1}{c|}{Common-}    & Micro &68.85 \cellcolor[HTML]{DAE8FC} &78.69 \cellcolor[HTML]{DAE8FC} &61.58 \cellcolor[HTML]{DAE8FC} &89.65 \cellcolor[HTML]{DAE8FC} &92.93 \cellcolor[HTML]{DAE8FC} \\ 
\Xcline{3-3}{0.4pt}
\multicolumn{1}{c|}{Test}       & \multicolumn{1}{c|}{sense}      & Macro &0.00 \cellcolor[HTML]{DAE8FC} &8.20 \cellcolor[HTML]{DAE8FC} &3.28 \cellcolor[HTML]{DAE8FC} &52.46 \cellcolor[HTML]{DAE8FC} &54.92 \cellcolor[HTML]{DAE8FC} \\ 
\Xcline{2-3}{0.4pt}
\multicolumn{1}{c|}{Set (122)}  & \multicolumn{1}{c|}{Hard}       & Micro &0.82 \cellcolor[HTML]{DAE8FC} &13.93 \cellcolor[HTML]{DAE8FC} &2.46 \cellcolor[HTML]{DAE8FC} &46.72 \cellcolor[HTML]{DAE8FC} &50.00 \cellcolor[HTML]{DAE8FC} \\ 
\Xcline{3-3}{0.4pt}
\multicolumn{1}{c|}{}           & \multicolumn{1}{c|}{Constraint} & Macro &0.82 \cellcolor[HTML]{DAE8FC} &13.93 \cellcolor[HTML]{DAE8FC} &2.46 \cellcolor[HTML]{DAE8FC} &46.72 \cellcolor[HTML]{DAE8FC} &50.00 \cellcolor[HTML]{DAE8FC} \\ 
\Xcline{2-3}{0.4pt}
\multicolumn{1}{c|}{}           & \multicolumn{2}{c|}{\colorr{Final Pass Rate}}    &0.00 \cellcolor[HTML]{DAE8FC} &4.10 \cellcolor[HTML]{DAE8FC} &1.64 \cellcolor[HTML]{DAE8FC} &\textbf{\colorb{43.44}} \cellcolor[HTML]{DAE8FC} &30.33 \cellcolor[HTML]{DAE8FC} \\ 
\Xhline{1.5pt}
\hline
\end{tabular}
}
\end{table*}

\begin{table*}[h]
\centering
\caption{The Pass Rates (\%) on \textbf{Medium} Instances. The highest final pass rates are highlighted in bold blue.}
\label{performance-on-medium-instances-TravelPlanner-with-commensense}
\setlength{\extrarowheight}{3pt}
\scalebox{0.7}{
\begin{tabular}{@{}ccc|cc|cc|c@{}}
\hline
\Xhline{1.5pt}
\multicolumn{3}{c|}{With Unconventional Hint}                             & \zc{GPT-3.5} + ReAct          & \zc{GPT-4} + ReAct          & \zc{GPT-3.5} + \zc{PMC}            & \zc{GPT-4} + \zc{PMC}            & \zc{GPT-4} (SP)             \\ 
\Xhline{1pt}
\multicolumn{1}{c|}{}           & \multicolumn{2}{c|}{Delivery Rate}      &100.00 \cellcolor[HTML]{EFEFEF} &100.00 \cellcolor[HTML]{EFEFEF} &100.00 \cellcolor[HTML]{EFEFEF} &95.00 \cellcolor[HTML]{EFEFEF} &100.00 \cellcolor[HTML]{EFEFEF} \\ 
\Xcline{2-3}{0.4pt}
\multicolumn{1}{c|}{}           & \multicolumn{1}{c|}{Common-}    & Micro &76.88 \cellcolor[HTML]{EFEFEF} &83.75 \cellcolor[HTML]{EFEFEF} &68.75 \cellcolor[HTML]{EFEFEF} &82.50 \cellcolor[HTML]{EFEFEF} &91.88 \cellcolor[HTML]{EFEFEF} \\ 
\Xcline{3-3}{0.4pt}
\multicolumn{1}{c|}{Validation} & \multicolumn{1}{c|}{sense}      & Macro &0.00 \cellcolor[HTML]{EFEFEF} &10.00 \cellcolor[HTML]{EFEFEF} &5.00 \cellcolor[HTML]{EFEFEF} &20.00 \cellcolor[HTML]{EFEFEF} &50.00 \cellcolor[HTML]{EFEFEF} \\ 
\Xcline{2-3}{0.4pt}
\multicolumn{1}{c|}{Set (20)}   & \multicolumn{1}{c|}{Hard}       & Micro &0.00 \cellcolor[HTML]{EFEFEF} &7.50 \cellcolor[HTML]{EFEFEF} &0.00 \cellcolor[HTML]{EFEFEF} &55.00 \cellcolor[HTML]{EFEFEF} &55.00 \cellcolor[HTML]{EFEFEF} \\ 
\Xcline{3-3}{0.4pt}
\multicolumn{1}{c|}{}           & \multicolumn{1}{c|}{Constraint} & Macro &0.00 \cellcolor[HTML]{EFEFEF} &0.00 \cellcolor[HTML]{EFEFEF} &0.00 \cellcolor[HTML]{EFEFEF} &55.00 \cellcolor[HTML]{EFEFEF} &20.00 \cellcolor[HTML]{EFEFEF} \\ 
\Xcline{2-3}{0.4pt}
\multicolumn{1}{c|}{}           & \multicolumn{2}{c|}{\colorr{Final Pass Rate}}    &0.00 \cellcolor[HTML]{EFEFEF} &0.00 \cellcolor[HTML]{EFEFEF} &0.00 \cellcolor[HTML]{EFEFEF} &\textbf{\colorb{15.00}} \cellcolor[HTML]{EFEFEF} &5.00 \cellcolor[HTML]{EFEFEF} \\ 
\Xhline{1.5pt}
\hline
\multicolumn{1}{c|}{}           & \multicolumn{2}{c|}{Delivery Rate}      &94.23 \cellcolor[HTML]{DAE8FC} &98.08 \cellcolor[HTML]{DAE8FC} &85.58 \cellcolor[HTML]{DAE8FC} &95.19 \cellcolor[HTML]{DAE8FC} &100.00 \cellcolor[HTML]{DAE8FC} \\ 
\Xcline{2-3}{0.4pt}
\multicolumn{1}{c|}{}           & \multicolumn{1}{c|}{Common-}    & Micro &70.67 \cellcolor[HTML]{DAE8FC} &78.97 \cellcolor[HTML]{DAE8FC} &59.86 \cellcolor[HTML]{DAE8FC} &88.46 \cellcolor[HTML]{DAE8FC} &89.78 \cellcolor[HTML]{DAE8FC} \\ 
\Xcline{3-3}{0.4pt}
\multicolumn{1}{c|}{Test}       & \multicolumn{1}{c|}{sense}      & Macro &0.00 \cellcolor[HTML]{DAE8FC} &9.62 \cellcolor[HTML]{DAE8FC} &0.00 \cellcolor[HTML]{DAE8FC} &50.96 \cellcolor[HTML]{DAE8FC} &39.42 \cellcolor[HTML]{DAE8FC} \\ 
\Xcline{2-3}{0.4pt}
\multicolumn{1}{c|}{Set (104)}  & \multicolumn{1}{c|}{Hard}       & Micro &0.00 \cellcolor[HTML]{DAE8FC} &12.50 \cellcolor[HTML]{DAE8FC} &0.96 \cellcolor[HTML]{DAE8FC} &50.48 \cellcolor[HTML]{DAE8FC} &48.08 \cellcolor[HTML]{DAE8FC} \\ 
\Xcline{3-3}{0.4pt}
\multicolumn{1}{c|}{}           & \multicolumn{1}{c|}{Constraint} & Macro &0.00 \cellcolor[HTML]{DAE8FC} &6.73 \cellcolor[HTML]{DAE8FC} &0.96 \cellcolor[HTML]{DAE8FC} &40.38 \cellcolor[HTML]{DAE8FC} &22.12 \cellcolor[HTML]{DAE8FC} \\ 
\Xcline{2-3}{0.4pt}
\multicolumn{1}{c|}{}           & \multicolumn{2}{c|}{\colorr{Final Pass Rate}}    &0.00 \cellcolor[HTML]{DAE8FC} &1.92 \cellcolor[HTML]{DAE8FC} &0.00 \cellcolor[HTML]{DAE8FC} &\textbf{\colorb{34.62}} \cellcolor[HTML]{DAE8FC} &5.77 \cellcolor[HTML]{DAE8FC} \\ 
\Xhline{1.5pt}
\hline
\end{tabular}
}
\end{table*}

\begin{table*}[h]
\centering
\caption{The Pass Rates (\%) on \textbf{Hard} Instances. The highest final pass rates are highlighted in bold blue.}
\label{performance-on-hard-instances-TravelPlanner-with-commensense}
\setlength{\extrarowheight}{3pt}
\scalebox{0.7}{
\begin{tabular}{@{}ccc|cc|cc|c@{}}
\hline
\Xhline{1.5pt}
\multicolumn{3}{c|}{With Unconventional Hint}                             & \zc{GPT-3.5} + ReAct          & \zc{GPT-4} + ReAct          & \zc{GPT-3.5} + \zc{PMC}            & \zc{GPT-4} + \zc{PMC}            & \zc{GPT-4} (SP)             \\ 
\Xhline{1pt}
\multicolumn{1}{c|}{}           & \multicolumn{2}{c|}{Delivery Rate}      &95.00 \cellcolor[HTML]{EFEFEF} &100.00 \cellcolor[HTML]{EFEFEF} &85.00 \cellcolor[HTML]{EFEFEF} &95.00 \cellcolor[HTML]{EFEFEF} &100.00 \cellcolor[HTML]{EFEFEF} \\ 
\Xcline{2-3}{0.4pt}
\multicolumn{1}{c|}{}           & \multicolumn{1}{c|}{Common-}    & Micro &69.34 \cellcolor[HTML]{EFEFEF} &79.38 \cellcolor[HTML]{EFEFEF} &60.00 \cellcolor[HTML]{EFEFEF} &83.75 \cellcolor[HTML]{EFEFEF} &89.38 \cellcolor[HTML]{EFEFEF} \\ 
\Xcline{3-3}{0.4pt}
\multicolumn{1}{c|}{Validation} & \multicolumn{1}{c|}{sense}      & Macro &0.00 \cellcolor[HTML]{EFEFEF} &10.00 \cellcolor[HTML]{EFEFEF} &0.00 \cellcolor[HTML]{EFEFEF} &40.00 \cellcolor[HTML]{EFEFEF} &35.00 \cellcolor[HTML]{EFEFEF} \\ 
\Xcline{2-3}{0.4pt}
\multicolumn{1}{c|}{Set (20)}   & \multicolumn{1}{c|}{Hard}       & Micro &1.25 \cellcolor[HTML]{EFEFEF} &5.00 \cellcolor[HTML]{EFEFEF} &0.00 \cellcolor[HTML]{EFEFEF} &41.25 \cellcolor[HTML]{EFEFEF} &50.00 \cellcolor[HTML]{EFEFEF} \\ 
\Xcline{3-3}{0.4pt}
\multicolumn{1}{c|}{}           & \multicolumn{1}{c|}{Constraint} & Macro &0.00 \cellcolor[HTML]{EFEFEF} &0.00 \cellcolor[HTML]{EFEFEF} &0.00 \cellcolor[HTML]{EFEFEF} &30.00 \cellcolor[HTML]{EFEFEF} &5.00 \cellcolor[HTML]{EFEFEF} \\ 
\Xcline{2-3}{0.4pt}
\multicolumn{1}{c|}{}           & \multicolumn{2}{c|}{\colorr{Final Pass Rate}}    &0.00 \cellcolor[HTML]{EFEFEF} &0.00 \cellcolor[HTML]{EFEFEF} &0.00 \cellcolor[HTML]{EFEFEF} &\textbf{\colorb{30.00}} \cellcolor[HTML]{EFEFEF} &0.00 \cellcolor[HTML]{EFEFEF} \\ 
\Xhline{1.5pt}
\hline
\multicolumn{1}{c|}{}           & \multicolumn{2}{c|}{Delivery Rate}      &97.56 \cellcolor[HTML]{DAE8FC} &97.56 \cellcolor[HTML]{DAE8FC} &73.17 \cellcolor[HTML]{DAE8FC} &100.00 \cellcolor[HTML]{DAE8FC} &100.00 \cellcolor[HTML]{DAE8FC} \\ 
\Xcline{2-3}{0.4pt}
\multicolumn{1}{c|}{}           & \multicolumn{1}{c|}{Common-}    & Micro &72.56 \cellcolor[HTML]{DAE8FC} &74.24 \cellcolor[HTML]{DAE8FC} &49.24 \cellcolor[HTML]{DAE8FC} &91.46 \cellcolor[HTML]{DAE8FC} &91.01 \cellcolor[HTML]{DAE8FC} \\ 
\Xcline{3-3}{0.4pt}
\multicolumn{1}{c|}{Test}       & \multicolumn{1}{c|}{sense}      & Macro &1.22 \cellcolor[HTML]{DAE8FC} &3.66 \cellcolor[HTML]{DAE8FC} &2.44 \cellcolor[HTML]{DAE8FC} &50.00 \cellcolor[HTML]{DAE8FC} &39.02 \cellcolor[HTML]{DAE8FC} \\ 
\Xcline{2-3}{0.4pt}
\multicolumn{1}{c|}{Set (82)}  & \multicolumn{1}{c|}{Hard}       & Micro &2.13 \cellcolor[HTML]{DAE8FC} &14.02 \cellcolor[HTML]{DAE8FC} &1.22 \cellcolor[HTML]{DAE8FC} &53.96 \cellcolor[HTML]{DAE8FC} &56.71 \cellcolor[HTML]{DAE8FC} \\ 
\Xcline{3-3}{0.4pt}
\multicolumn{1}{c|}{}           & \multicolumn{1}{c|}{Constraint} & Macro &0.00 \cellcolor[HTML]{DAE8FC} &7.32 \cellcolor[HTML]{DAE8FC} &0.00 \cellcolor[HTML]{DAE8FC} &45.12 \cellcolor[HTML]{DAE8FC} &7.32 \cellcolor[HTML]{DAE8FC} \\ 
\Xcline{2-3}{0.4pt}
\multicolumn{1}{c|}{}           & \multicolumn{2}{c|}{\colorr{Final Pass Rate}}    &0.00 \cellcolor[HTML]{DAE8FC} &2.44 \cellcolor[HTML]{DAE8FC} &0.00 \cellcolor[HTML]{DAE8FC} &\textbf{\colorb{42.68}} \cellcolor[HTML]{DAE8FC} &3.66 \cellcolor[HTML]{DAE8FC} \\ 
\Xhline{1.5pt}
\hline
\end{tabular}
}
\end{table*}


\section{\gxd{Prompt Optimization for Each Agent}}\label{appendix:prompt-optimization}

\gxd{We perform the prompt optimization for each agent by observing the performance of agents in a few samples from the validation datasets. Once the agent achieves a fairly well result, we use the same prompt for the agent on the test dataset and do not further optimize the prompts.}

\subsection{\gxd{General Prompt Optimizations Across All Agents}}
\gxd{There are mainly two types of problems that needed to be addressed across all agents via prompt optimizations. (1) Hallucinations in agents. To tackle such problems, our prompts will inform the agents to understand what information they have, e.g. retrieved information and tools. We also guide the agents to explain their reasoning via Chain-of-Thoughts (CoT) before providing the required output. (2) The specific requirement from the benchmark. Our benchmarks have specific requirements that are not generalisable on other use cases. For example, TravelPlanner only takes one global constraint into account for evaluation despite there are many other global constraints in actual trip planning. The benchmark also requires the final output to be delivered in a structured format for evaluation. We added additional instructions for the agents to meet such requirements. However, we don’t provide direct answers in our prompt to maintain the zero-shot property.} 

\gxd{The Manager Agent, Executor Agents, Supervisor Agent and Deliverer Agent require additional prompt optimization work due to the additional problems the agents may face. The optimization for each agent is elaborated separately in the following sections.} 

\subsection{\gxd{The Manager Agent}}
\gxd{The design of prompt for the manager agent requires more effort in comparison to other agents due to the challenging nature of agents performing a complicated planning task in zero-shot. We observe that most of the available language models in our experiments are able to provide output based on a desired structure. Based on the task planning in PMC (refer to ~\ref{hierarchical}), we design a corresponding JSON template consisting of subtasks and constraints for the manager agent to fill. The JSON will provide the information needed to automated the subsequent workflow. The structure of the JSON template is designed in a way that can be parsed by Python code to automatically execute the PMC workflow.}

\begin{figure*}[!ht]
    \centering
    {\includegraphics[width=1\textwidth]{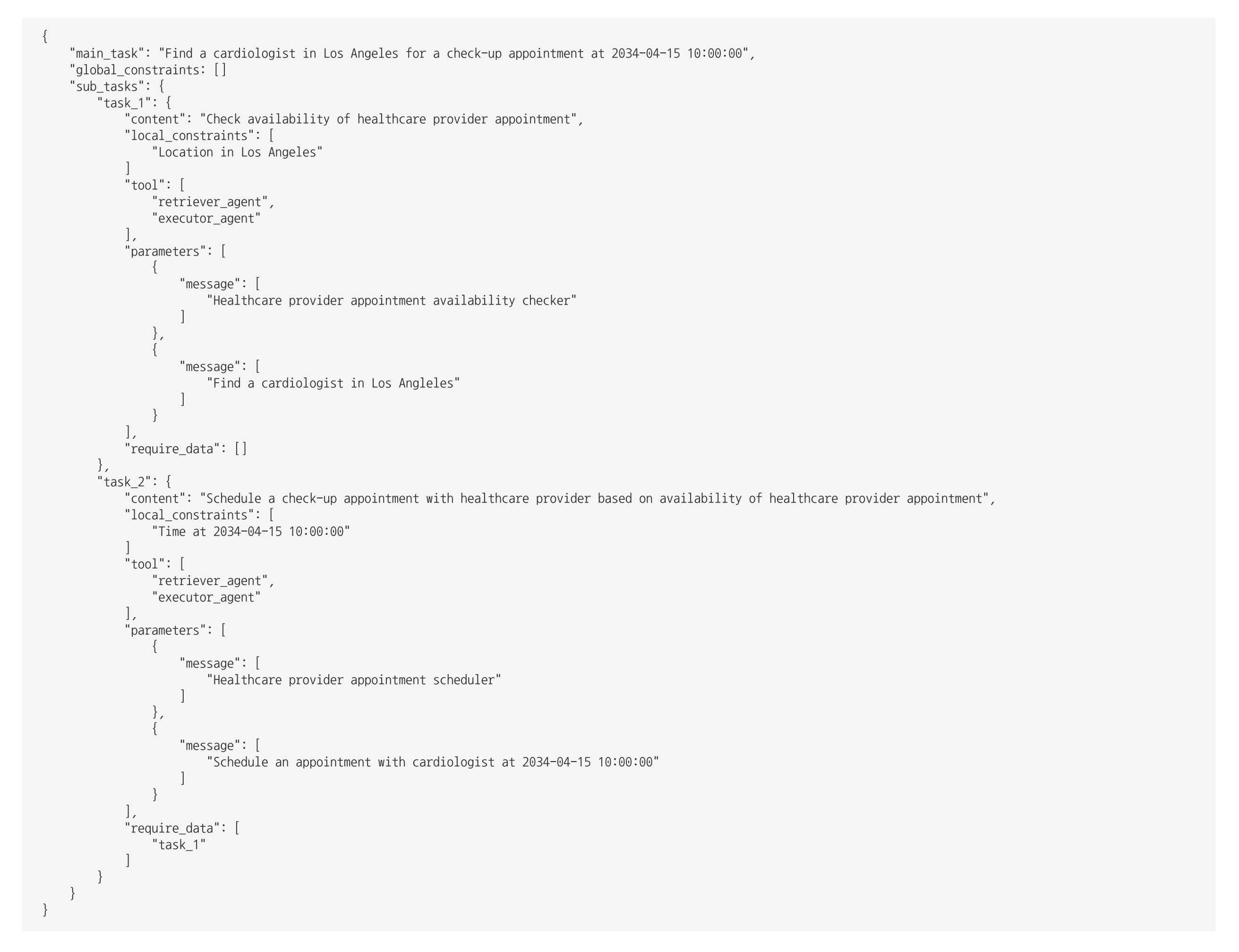}}
    \caption{Step 1 of prompt optimization for Manager Agent. Full demonstration provided. This demonstration above consists of two sub-tasks, the number of sub-tasks will vary based on the main task. The conversion of prompt into zero-shot is shown in Figure ~\ref{fig:demo-2}, ~\ref{fig:demo-3} and ~\ref{fig:demo-4}.}
    \label{fig:demo-1}
\end{figure*}

\begin{figure*}[!ht]
    \centering
    {\includegraphics[width=1\textwidth]{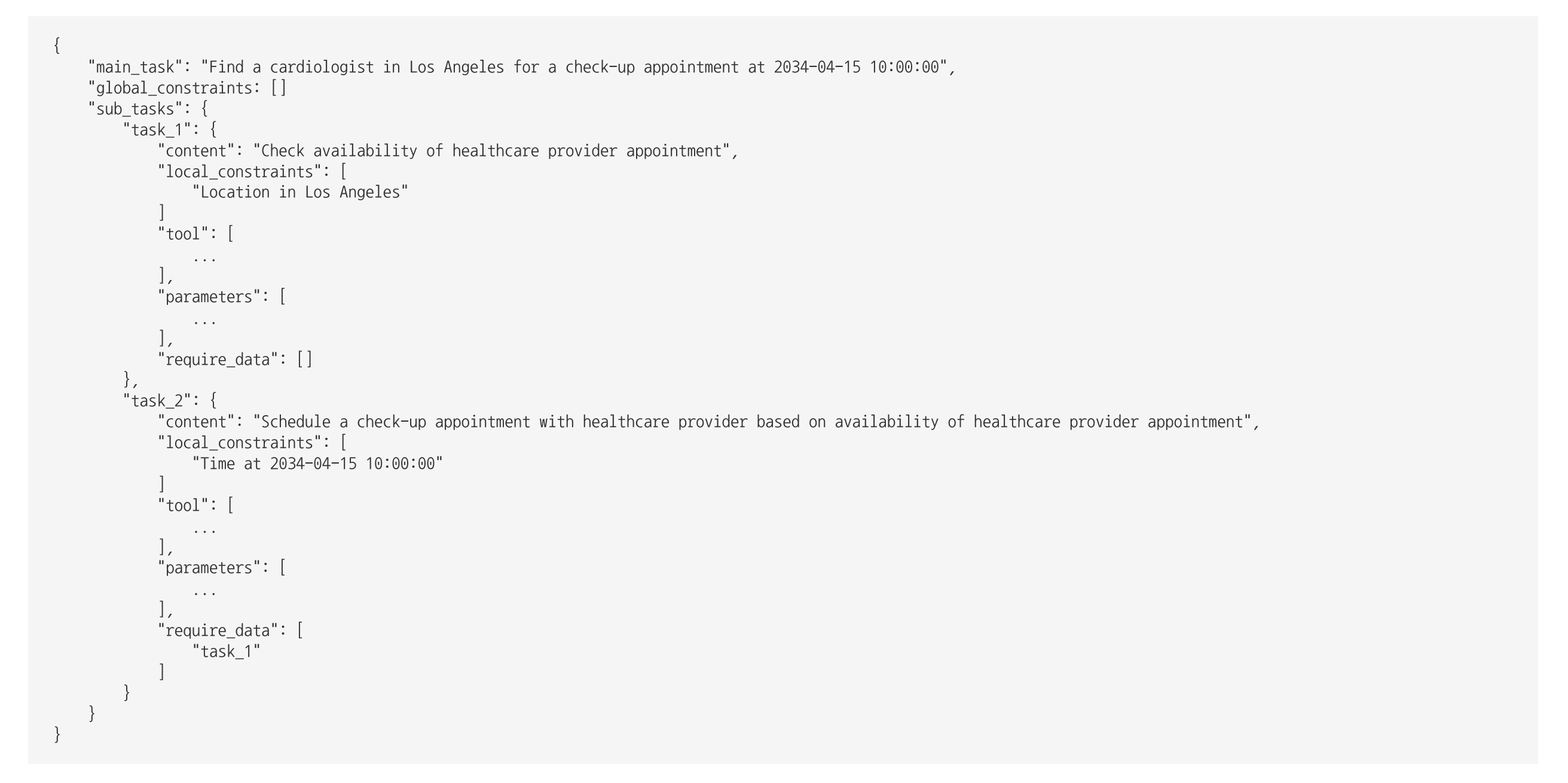}}
    \caption{Step 2 of prompt optimization for Manager Agent. The demonstration of filling tools and parameters values for each task are removed. This demonstration above consists of two sub-tasks, the number of sub-tasks will vary based on the main task.}
    \label{fig:demo-2}
\end{figure*}

\begin{figure*}[!ht]
    \centering
    {\includegraphics[width=1\textwidth]{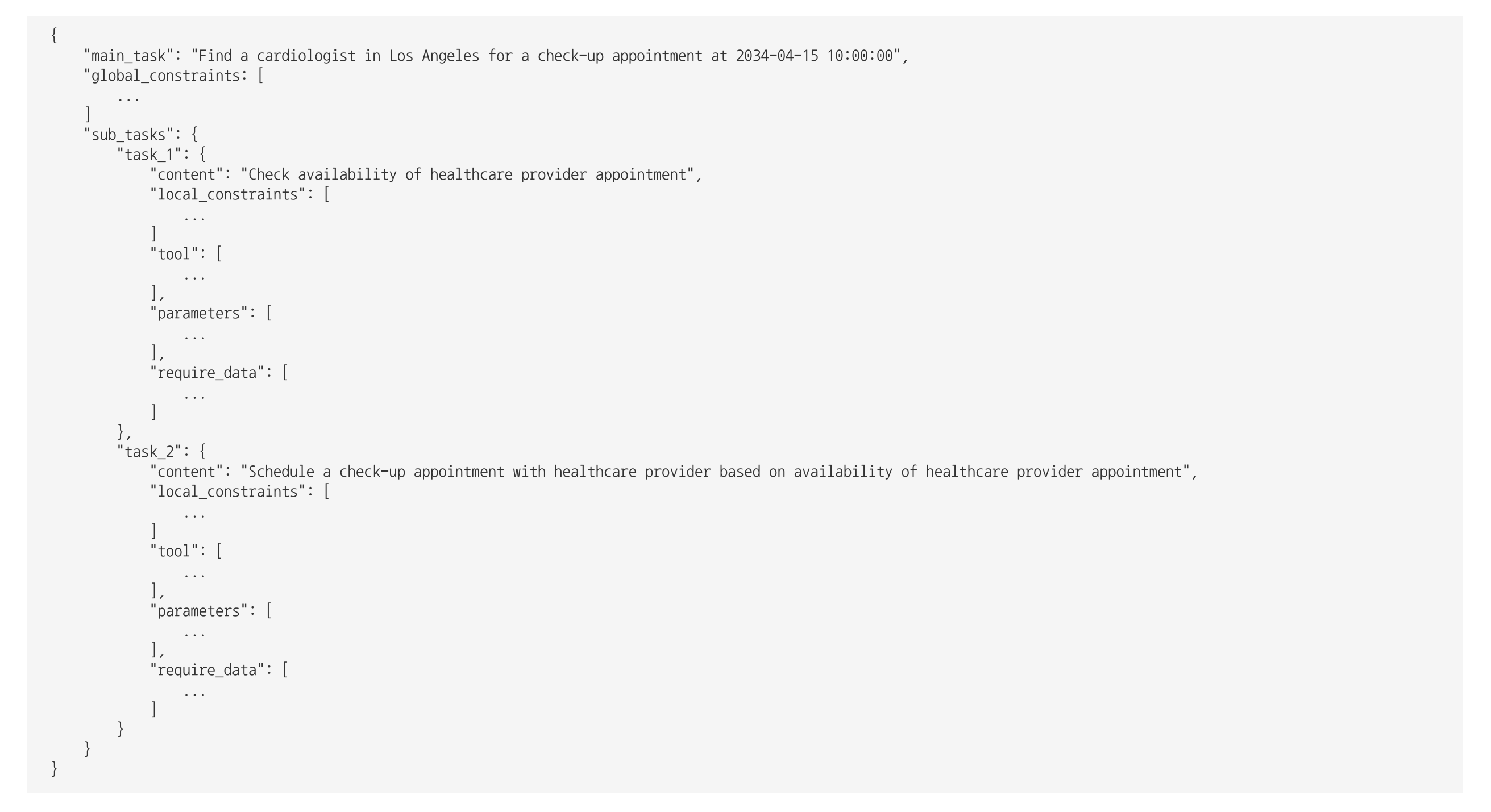}}
    \caption{Step 3 of prompt optimization for Manager Agent. The demonstration of filling constraints and require data (task dependency) values for each task are removed. This demonstration above consists of two sub-tasks, the number of sub-tasks will vary based on the main task.}
    \label{fig:demo-3}
\end{figure*}

\begin{figure*}[!ht]
    \centering
    {\includegraphics[width=1\textwidth]{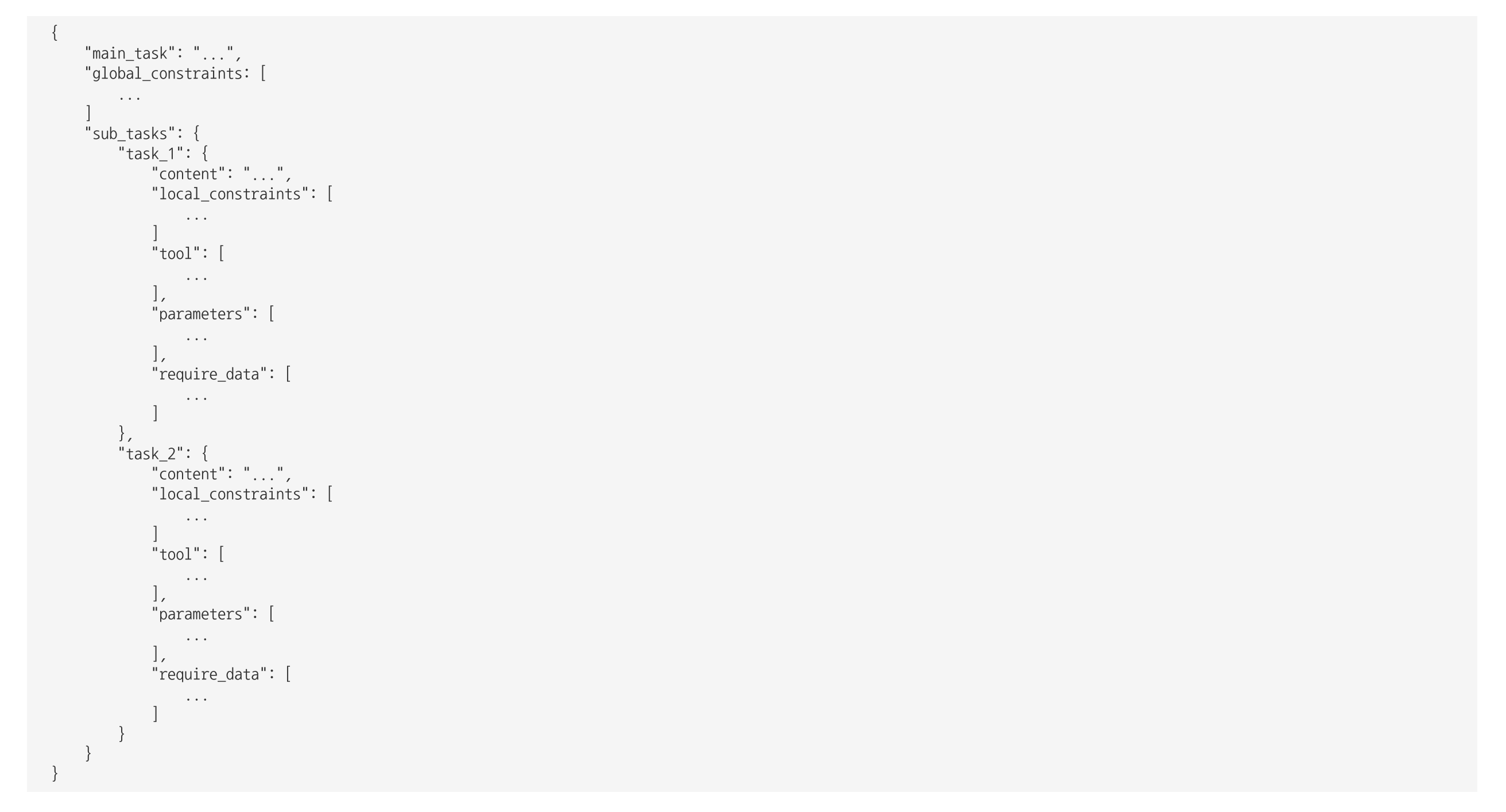}}
    \caption{Step 4 of prompt optimization for Manager Agent. The demonstration of filling main task and sub-task description are not provided. This demonstration above consists of two sub-tasks, the number of sub-tasks will vary based on the main task.}
    \label{fig:demo-4}
\end{figure*}

\gxd{We first select five prompts from the validation dataset and provide few-shot examples in the demonstrations for the agent to observe its planning. The agent is usually capable of producing correct plans from unseen queries given such few-shot demonstrations. We replace one or more elements from each example, e.g. tools, parameters and constraints with ellipsis. The ellipsis indicates that no demonstration is given on how to fill the value for the element. We add additional instructions to the prompt to show how to fill the values, and optimize the instructions until the manager produces the desired plan. Then, we replace other elements from the example with ellipsis and continue adding instructions to the prompt, until all values are ellipsis in the JSON template (refer to Figure ~\ref{fig:demo-4}). Hence, the final prompt for the agent is a zero-shot prompt. Figure ~\ref{fig:demo-1}, ~\ref{fig:demo-2}, ~\ref{fig:demo-3} and ~\ref{fig:demo-4} shows the step-by-step iteration of how we optimize the prompt from few-shot demonstration to zero-shot.}

\subsection{\gxd{The Executor Agent}}
\gxd{The design of prompt for the executor agent is relatively easy as the executor agent is only required to execute step-level tasks like commonly seen tool agents. However, based on our observation and paper ~\cite{xie2024travelplanner}, there are some common problems that the language models may suffer in tool execution. The model may repeatedly use the same tools with same parameters and not fulfilling the task. The problems can be mitigated by adding some rules in the prompt. To avoid repeated tool usage, the agent should mention the current information it has obtained so far and the next information it requires. As Executor Agent might have multiple tools, when the current tool could not meet the task objective, the agent should also attempt to use other tools instead of giving up. Since the task outcome by the executor agent may be delivered to the Supervisor Agent and Deliverer Agent, the executor agent should also be instructed to provide very detailed information with clear structures to ensure high readability by other agents without any loss of information.}

\subsection{\gxd{The Supervisor Agent}}
\gxd{The design of prompt for the supervisor agent is relatively easy as task rewriting is not complicated. Our prompt requires the supervisor agent to rewrite the task with the similar JSON format we gave to the manager agent.}

\subsection{\gxd{The Deliverer Agent}}
\gxd{The design of prompt for the deliverer agent requires a fair amount of effort as different task, e.g, TravelPlanner and APIBank, requires the Deliverer Agent to perform different primary role. Hence, our prompt has to be adjusted to meet the primary role. We believe the primary role of deliverer agent can be divided into two types, "Inference Role" and "Reporting Role". In TravelPlanner, the Deliverer Agent is responsible for the "Inference Role", which its job is to deliver a plan given the execution results following the same format as experiments in the original paper ~\cite{xie2024travelplanner}. The "Inference Role" requires additional thinking from the agent as the agent needs to make further inference based on the acquired subtask results and plan requires to meet the global constraints of the task. For such role, we discover the agent is aware of the global constraints, but it may forget some commonsense that might needed to be considered in its inference. Hence, our prompt requires to remind the agent to be aware of commonsense when delivering its plan. Unlike the "Inference Role", the "Reporting Role" does not require to make any further plan, but only report or summarize the outcome of all subtasks. Based on the requirement of the task, our prompt will mention the level of details the final response should contain.}

\gxd{Despite it sounds effortful to redesign a prompt for each task, however in most scenario, the primary role of the Deliverer Agent falls into either one of these two categories. Hence, given a new task, we only need to identify the primary role of Deliverer Agent, and reuse one of the two prompts, with little modification based on specific rules of the task.}

\section{\zc{Prompt and Instructions for Each Agent}}\label{appendix:prompt-design}
\zc{We give the prompt structure for each agent in Figure~\ref{fig:agent-design}.}

\lstset{style=mystyle}

\subsection{TravelPlanner}\label{appendix:prompt-design-TravelPlanner}

\textbf{The Manager Agent}:\label{appendix:prompt-design-manager-TravelPlanner}
\begin{lstlisting}[breakindent=0pt]
You are a task management assistant designed to break down tasks and manage task progress. 

The main job in task breakdown is populating the JSON template below:
```json
    {   
    "main_task": "...",
    "global_constraints": [...],
    "sub_tasks": {
        "task_1": {"content": "...", "tool": [...], "parameters": [{...}], "local_constraints": [...], "require_data": [...]},
        "task_2": {"content": "...", "tool": [...], "parameters": [{...}], "local_constraints": [...], "require_data": [...]}
        }
    }
```

Based on user's query, your main task is to gather valid information related to transportation, dining, attractions, and accommodation using the capabilities of tools.

Before you design your task, you should understand what tools you have, what each tool can do and cannot do. You must not design the subtask that do not have suitable tool to perform. Never design subtask that does not use any tool.

You must first output the Chain of Thoughts (COT). In the COT, you need to explain how you break down the main task into sub-tasks and justify why each subtask can be completed by a tool. The sub-tasks need to be broken down to a very low granularity, hence it's possible that some sub-tasks will depend on the execution results of previous tasks. You also need to specify which sub-tasks require the execution results of previous tasks. When writing about each sub-task, you must also write out its respective local constraints. Finally, you write the global constraint of the main task.

All the results of the sub-tasks will be passed to the "interactor_agent". The "interactor_agent" has various capabilities such as inference, computation, and generating responses, but it cannot be used to answer unknown questions. You don't need to specify in the JSON template to call the "interactor_agent".

Before filling in the template, you must first understand the user's request, carefully analyzing the tasks contained within it. Once you have a clear understanding of the tasks, you determine the sequence in which each task should be executed. Following this sequence, you rewrite the tasks into complete descriptions, taking into account the dependencies between them.

In the JSON template you will be filling, "main_task" is your main task, which is gather valid information related to transportation by flight and car, dining, attractions, and accommodation based on user's query. "sub_task" is the sub-tasks that you would like to break down the task into. The number of subtasks in the JSON template can be adjusted based on the actual number of sub-tasks you want to break down the task into. The break down process of the sub-tasks must be simple with low granularity. There is no limit to the number of subtasks.
Each sub-tasks consist of either one or multiple step. It contains 5 information to be filled in, which are "content", "agent", "parameters", "require_data" and "data".

"require_data" is a list of previous sub-tasks which their information is required by the current sub-task. Some sub-tasks require the information of previous sub-task. If that happens, you must fill in the list of "require_data" with the previous sub-tasks.

"content" is the description of the subtask, formatted as string. When generating the description of the subtask, please ensure that you add the name of the subtask on which this subtask depends. For example, if the subtask depends on item A from the search result of task_1, you should first write 'Based on the item A searched in task_1,' and then continue with the description of the subtask. It is important to indicate the names of the dependent subtasks.

"tool" is the list of tools required for each step of execution. Please use the original name of the tool without "functions." in front. This list cannot be empty. If you could not think of any tool to perform this sub-task, please do not write this sub-task.

"parameters" is a list specifying the parameters required for each tool. Within the "parameters" list, the format for "message" is list, which will be concatenated by the system and passed to the tool.

After determining your subtasks, you must first identify the local constraints for each sub-task, then the global constraints. Local constraints are constraints that needed to be considered in only in each specific sub-task. Meanwhile, global constraints are the constraints mentioned in the query that needed to be jointly considered across all the sub-tasks. You must not write global constraints that are only related to only some of the sub-tasks.

Please write the local constraints of each sub-task in its corresponding "local_constraints" and the write the global constraints in "global_constraints". You should not write global constraints into "local_constraints". Similarly, constraints that exists in "local_constraints" should not be written in "global_constraints". Hence, local constraints of each sub-task and global constraints must be unique. You don't need to design a task specifically for global constraints as the global constraints will be passed to the "interactor_agent".
When writing "local_constraints", please write it as specific as possible, as you should assume the tools of each task have no knowledge of the user's query. You should also be aware local constraints filters the items individually, and some constraints can only be satisfied by multiple items. For example, if user's constraints is "Indian, Chinese and Mediterranean cuisine", your local constraints should be "Indian, Chinese or Mediterranean cuisine" instead of "Indian, Chinese and Mediterranean cuisine" because it is not possible for a single restaurant to have all cuisine but multiple restaurants that meet one types of cuisine can be combined to meet the constraint.

Important Reminder : Global constraint is constraint that are jointly considered across all the sub-tasks. You must not write global constraint that is only related to only some of the sub-tasks. For example, a constraint related to accomodation only should not be considered as "global constraints" as there are tasks unrelated to accomodation. In your COT, please justify why your global_constraints shuld be considered by all of the sub-tasks.

Important Rule : You must only output one global constraint that you think is the most important one based on the user query.

You must output the JSON at the end.
\end{lstlisting}

\textbf{The Manager Agent (LLaMA-3.1-8B)}:\label{appendix:prompt-design-managerllama-TravelPlanner}

\begin{lstlisting}[breakindent=0pt]
You are a task management assistant designed to break down tasks and manage task progress. 

The main job in task breakdown is populating the JSON template below:
{   
    "global_constraints": [...],
    "sub_tasks": {
        "task_1": {"content": "...", "tool": [...], "parameters": [{"message": ["..."]}], "local_constraints": [...], "require_data": [...]},
        "task_2": {"content": "...", "tool": [...], "parameters": [{"message": ["..."]}], "local_constraints": [...], "require_data": [...]}
        }
}

Based on user's query, your main task is to gather valid information related to transportation, accomodation, dining and attractions using the capabilities of tools.

When filling in local_constraints, you must fill in the constraints based on the user requirements.

Here is an example of user query with some local constraints mentioned by user and example output.

User Query :
Could you organize a 3-day trip for 3 people from Raleigh to Boise, spanning from January 1st to January 3rd, 2022, with a budget of $2,266?

Output :
[START]
{
    "global_constraints": ["Total budget must not exceed $2266"],
    "main_task": "Plan a trip from Raleigh to Boise for 3 days, from January 1st to January 3rd, 2022, within a budget of $2,266",
    "sub_tasks": {
        "task_1": {
            "content": "Find round-trip transportation options from Raleigh to Boise on January 1st, 2022, and returning on January 3rd, 2022",
            "tool": ["search_cross_city_transport_agent"],
            "parameters": [{
                "message": ["Flight from Raleigh to Boise on 2022-01-01 and Boise to Raleigh on 2022-01-03."]
            }],
            "local_constraints": [],
            "require_data": []
        },
        "task_2": {
            "content": "Search for accommodations in Boise for 2 nights, from January 1st to January 3rd, 2022",
            "tool": ["search_city_accommodation_agent"],
            "parameters": [{
                "message": ["Accommodations in Boise for maximum of 2 nights."]
            }],
            "local_constraints": [],
            "require_data": []
        },
        "task_3": {
            "content": "Search for restaurants in Boise",
            "tool": ["search_city_hospitality_agent"],
            "parameters": [{
                "message": ["Restaurants in Boise."]
            }],
            "local_constraints": [],
            "require_data": []
        },
        "task_4": {
            "content": "Search for attractions in Boise",
            "tool": ["search_city_hospitality_agent"],
            "parameters": [{
                "message": ["Attractions in Boise."]
            }],
            "local_constraints": [],
            "require_data": []
        }
    }
}
[END]

Please do not fill in local_constraints that user does not mention. For example, if user does not say "no self-driving" as transportation requirement, please do not include that in local_constraints.
You only need to output the JSON. Please wrap your JSON with [START] and [END]. Do not output anything else. Do not write any Python script.
\end{lstlisting}

\textbf{The List of Executors for The Manager Agent}:\label{appendix:function-call-manager-TravelPlanner}

\begin{lstlisting}[breakindent=0pt]
[
    {'name': 'search_cross_city_transport_agent', 'description': "It can search details and availability of transportations from one city to another given a date. Please include the year of date. The available transportations are 'Flight', 'Self-driving' and 'Taxi'. Based on user's query, please analyse whether it is one-way travel or round-trip. Second, please consider as many transportations as possible unless user specifies particular transportations to travel. You can search multiple transportations in a single query. Remember, it cannot search for transportation within the city. Don't use it to search something else.", 'parameters': {'type': 'object', 'properties': {'message': {'type': 'string', 'description': "Search query. Please write in list form. Please provide the full date including year. One way example : ['<All possible transportations> from <origin city> to <destination city> on <departing full date including year>.']. Round-trip example: ['<All possible transportations> from <origin city> to <destination city> on <departing full date including year> and <destination city> to <origin city> on <returning full date including year>.']"}}, 'required': ['message']}}, 

    {'name': 'search_city_hospitality_agent', 'description': "It can search details of hospitality of a city. The available hospitality are 'Attractions' and 'Restaurants'. You can search multiple hospitality in a single query. Don't use it to search something else.", 'parameters': {'type': 'object', 'properties': {'message': {'type': 'string', 'description': "Search query. Please write in list form. For example: ['<Hospitalities> in <city name>.']"}}, 'required': ['message']}}, 

    {'name': 'search_cities_in_state_agent', 'description': "It can search all the cities in a given state. Don't use it to search something else.", 'parameters': {'type': 'object', 'properties': {'message': {'type': 'string', 'description': "Search query. Please write in list form. For example: ['Find cities in <state name>']"}}, 'required': ['message']}}, 

    {'name': 'search_city_accommodation_agent', 'description': 'It can search details of accommodation of a city. You must determine how many nights are required for the stay. If it is a round trip for n days, please consider the nights required are n-1 days.', 'parameters': {'type': 'object', 'properties': {'message': {'type': 'string', 'description': "Search query. Please write in list form. For example: ['Accomodations in <city name> for maximum of <number of nights> nights.']"}}, 'required': ['message']}}
]
\end{lstlisting}

\textbf{The Executor Agents}:\label{appendix:prompt-design-executor-TravelPlanner}
\begin{lstlisting}[breakindent=0pt]
<@\textcolor{blue}{\# Prompt for Search Cities In State Agent}@>
You are a search agent, you can use different tools to search for the cities in a given state based on user's query. You can assume the tools work fine.

These are the rules you should follow:
1. Before you use a tool, you must output your reasoning of using the tool. You must mention what information you have obtained from previous use of tools and what information you are looking to obtain from the next use of tool.
2. If you cannot provide an informative response based on user query, please consider using alternative tools to provide alternative information.
3. Please do not make any assumptions using your internal knowledge.
4. After you gather all the information you need, please output the information based on user's query. Your information must be as detailed as possible.
5. You should only provide informative response based on user query. Don't provide any other advice.

For each item in your search result, you need to ensure you write out all the features. If the input you receive is each item with features in bullet point form, you must ensure all bullet points are listed. If the input you receive is CSV, you must ensure each column represent each feature of the item. Do not miss any detail of every feature.
Your output format is as below. If you use the tool multiple times, you must output multiple sets of the search result format below:

Search Result of <Type of Items>:
1. Name : <Name of Item 1> 
<Feature 1>: <Detail of Feature 1 of Item 1>
<Feature 2>: <Detail of Feature 2 of Item 1>
...
<Feature n>: <Detail of Feature n of Item 1>

2. Name : <Name of Item 2> 
<Feature 1>: <Detail of Feature 1 of Item 2>
<Feature 2>: <Detail of Feature 2 of Item 2>
...
<Feature n>: <Detail of Feature n of Item 2>

.
.
.

N. Name : <Name of Item N> 
<Feature 1>: <Detail of Feature 1 of Item N>
<Feature 2>: <Detail of Feature 2 of Item N>
...
<Feature n>: <Detail of Feature n of Item N>
\end{lstlisting}

\begin{lstlisting}[breakindent=0pt]
<@\textcolor{blue}{\# The Function List for Search Cities In State Agent}@>
[
    {'name': 'CitySearch', 'description': 'Find cities in a state of your choice.', 'parameters': {'type': 'object', 'properties': {'State': {'type': 'string', 'description': "The name of the state where you're seeking cities"}}, 'required': ['State']}}
]
\end{lstlisting}

\begin{lstlisting}[breakindent=0pt]
<@\textcolor{blue}{\# Prompt for Search City Accommodation Agent}@>
You are a search agent, you can use different tools to search for the information of accomodations, restaurants and attractions in a given city based on user's query. You can assume the tools work fine.

These are the rules you should follow:
1. Before you use a tool, you must output your reasoning of using the tool. You must mention what information you have obtained from previous use of tools and what information you are looking to obtain from the next use of tool.
2. If you cannot provide an informative response based on user query, please consider using alternative tools to provide alternative information.
3. Please do not make any assumptions using your internal knowledge.
4. After you gather all the information you need, please output the information based on user's query. Your information must be as detailed as possible.
5. You should only provide informative response based on user query. Don't provide any other advice.

For each item in your search result, you need to ensure you write out all the features. If the input you receive is each item with features in bullet point form, you must ensure all bullet points are listed. If the input you receive is CSV, you must ensure each column represent each feature of the item. Do not miss any detail of every feature.
Your output format is as below. If you use the tool multiple times, you must output multiple sets of the search result format below:

Search Result of <Type of Items>:
1. Name : <Name of Item 1> 
<Feature 1>: <Detail of Feature 1 of Item 1>
<Feature 2>: <Detail of Feature 2 of Item 1>
...
<Feature n>: <Detail of Feature n of Item 1>

2. Name : <Name of Item 2> 
<Feature 1>: <Detail of Feature 1 of Item 2>
<Feature 2>: <Detail of Feature 2 of Item 2>
...
<Feature n>: <Detail of Feature n of Item 2>

.
.
.

N. Name : <Name of Item N> 
<Feature 1>: <Detail of Feature 1 of Item N>
<Feature 2>: <Detail of Feature 2 of Item N>
...
<Feature n>: <Detail of Feature n of Item N>
\end{lstlisting}

\begin{lstlisting}[breakindent=0pt]
<@\textcolor{blue}{\# The Function List for Search City Accommodation Agent}@>
[
    {'name': 'AccomodationSearch', 'description': 'Discover accommodations in your desired city.', 'parameters': {'type': 'object', 'properties': {'City': {'type': 'string', 'description': "The name of the city where you're seeking accommodation."}}, 'required': ['City']}}
]
\end{lstlisting}

\begin{lstlisting}[breakindent=0pt]
<@\textcolor{blue}{\# Prompt for Search City Hospitality Agent}@>
You are a search agent, you can use different tools to search for the information of restaurants and attractions in a given city based on user's query. You can assume the tools work fine.

These are the rules you should follow:
1. Before you use a tool, you must output your reasoning of using the tool. You must mention what information you have obtained from previous use of tools and what information you are looking to obtain from the next use of tool.
2. If you cannot provide an informative response based on user query, please consider using alternative tools to provide alternative information.
3. Please do not make any assumptions using your internal knowledge.
4. After you gather all the information you need, please output the information based on user's query. Your information must be as detailed as possible.
5. You should only provide informative response based on user query. Don't provide any other advice.

For each item in your search result, you need to ensure you write out all the features. If the input you receive is each item with features in bullet point form, you must ensure all bullet points are listed. If the input you receive is CSV, you must ensure each column represent each feature of the item. Do not miss any detail of every feature.
Your output format is as below. If you use the tool multiple times, you must output multiple sets of the search result format below:

Search Result of <Type of Items>:
1. Name : <Name of Item 1> 
<Feature 1>: <Detail of Feature 1 of Item 1>
<Feature 2>: <Detail of Feature 2 of Item 1>
...
<Feature n>: <Detail of Feature n of Item 1>

2. Name : <Name of Item 2> 
<Feature 1>: <Detail of Feature 1 of Item 2>
<Feature 2>: <Detail of Feature 2 of Item 2>
...
<Feature n>: <Detail of Feature n of Item 2>

.
.
.

N. Name : <Name of Item N> 
<Feature 1>: <Detail of Feature 1 of Item N>
<Feature 2>: <Detail of Feature 2 of Item N>
...
<Feature n>: <Detail of Feature n of Item N>
\end{lstlisting}

\begin{lstlisting}[breakindent=0pt]
<@\textcolor{blue}{\# The Function List for Search City Hospitality Agent}@>
[
    {'name': 'AttractionSearch', 'description': 'Find attractions in a city of your choice.', 'parameters': {'type': 'object', 'properties': {'City': {'type': 'string', 'description': "The name of the city where you're seeking restaurants."}}, 'required': ['City']}}, 
    
    {'name': 'RestaurantSearch', 'description': 'Explore dining options in a city of your choice.', 'parameters': {'type': 'object', 'properties': {'City': {'type': 'string', 'description': "The name of the city where you're seeking restaurants."}}, 'required': ['City']}}
]
\end{lstlisting}

\begin{lstlisting}[breakindent=0pt]
<@\textcolor{blue}{\# Prompt for Search Cross City Transport Agent}@>
You are a search agent, you can use different tools to search for the information of flights, self-driving or taxi in a given city based on user's query. If the query mentions all possible transportation, you should search all three options. You can assume the tools work fine.

These are the rules you should follow:
1. Before you use a tool, you must output your reasoning of using the tool. You must mention what information you have obtained from previous use of tools and what information you are looking to obtain from the next use of tool.
2. If it is a round-trip, you should perform search for both ways for the same transportation.
3. If you cannot provide an informative response based on user query, please consider using alternative tools to provide alternative information.
4. Please do not make any assumptions using your internal knowledge.
5. After you gather all the information you need, please output the information based on user's query. Your information must be as detailed as possible.
6. You should only provide informative response based on user query. Don't provide any other advice.

For each item in your search result, you need to ensure you write out all the features. If the input you receive is each item with features in bullet point form, you must ensure all bullet points are listed. If the input you receive is CSV, you must ensure each column represent each feature of the item. Do not miss any detail of every feature.
Your output format is as below. If you use the tool multiple times, you must output multiple sets of the search result format below:

Search Result of <Type of Items>:
1. Name : <Name of Item 1> 
<Feature 1>: <Detail of Feature 1 of Item 1>
<Feature 2>: <Detail of Feature 2 of Item 1>
...
<Feature n>: <Detail of Feature n of Item 1>

2. Name : <Name of Item 2> 
<Feature 1>: <Detail of Feature 1 of Item 2>
<Feature 2>: <Detail of Feature 2 of Item 2>
...
<Feature n>: <Detail of Feature n of Item 2>

.
.
.

N. Name : <Name of Item N> 
<Feature 1>: <Detail of Feature 1 of Item N>
<Feature 2>: <Detail of Feature 2 of Item N>
...
<Feature n>: <Detail of Feature n of Item N>
\end{lstlisting}

\begin{lstlisting}[breakindent=0pt]
<@\textcolor{blue}{\# The Function List for Search Cross City Transport Agent}@>
[
    {'name': 'FlightSearch', 'description': 'A flight information retrieval tool. Example: FlightSearch[New York, London, 2022-10-01] would fetch flights from New York to London on October 1, 2022.', 'parameters': {'type': 'object', 'properties': {'Departure City': {'type': 'string', 'description': "The city you'll be flying out from."}, 'Destination City': {'type': 'string', 'description': 'The city you aim to reach.'}, 'Date': {'type': 'string', 'description': 'The date of your travel in YYYY-MM-DD format.'}}, 'required': ['Departure City', 'Destination City', 'Date']}}, 
    
    {'name': 'GoogleDistanceMatrix', 'description': 'Estimate the distance, time and cost between two cities.', 'parameters': {'type': 'object', 'properties': {'Origin': {'type': 'string', 'description': 'The departure city of your journey.'}, 'Destination': {'type': 'string', 'description': 'The destination city of your journey.'}, 'Mode': {'type': 'string', 'description': "The method of transportation. Choices include 'self-driving' and 'taxi'."}}, 'required': ['Origin', 'Destination', 'Mode']}}
]
\end{lstlisting}

\textbf{The Supervisor Agent}:\label{appendix:prompt-design-supervisor-TravelPlanner}
\begin{lstlisting}[breakindent=0pt]
You are a task rewriting assistant, responsible for rewriting tasks to simplify the process of executing them.

The main job of rewriting tasks involves rewriting this template:
{
    "content": "...",
    "tool": "[...]",
    "parameters": "[{...}]",
}

"content" is the description of the sub-task, the format is string.
"tool" is the required tool for the corresponding task, the format is list.
"parameters" is the required parameters for each step for the respective agent, the format is list. Within "parameters", the format of "message" is string, which the system will concatenate the elements of the list and pass them to the agent.

User will provide a task based on this template. This task can be simplified with additional information, which is now obtained through the execution of previous tasks.
Hence, based on the information from the previous task, you need to rewrite the content and parameters to simplify the task further. Remember, please do not modify other content.
You are also given some information about the main query from user, which may provide additional info to help rewriting the task.
\end{lstlisting}

\textbf{The Deliverer Agent}:\label{appendix:prompt-design-deliverer-TravelPlanner}

\begin{lstlisting}[breakindent=0pt]
You are a proficient planner. Based on the provided items and query, please give me a detailed plan, including specifics such as flight numbers (e.g., F0123456), restaurant names, and accommodation names. Note that all the information in your plan should be derived from the provided data. You must adhere to the format given in the example. Additionally, all details should align with commonsense. 
The provided items for each task are ranked in preferences order, from highest to lowest. Please prioritise the higher ranking options in your plan but also make sure meet all the constraints from the query.
The symbol '-' indicates that information is unnecessary. For example, in the provided sample, you do not need to plan after returning to the departure city. When you travel to two cities in one day, you should note it in the 'Current City' section as in the example (i.e., from A to B).
Before you write your detailed plan, please analyse the hard constraints based on the query. In addition to that, we will also give you the hard constraints that we have analysed so far from the query. You also need to analyse the commonsense constraints for a diverse and sensible trip plan. Your commonsense constraints must also include not repeating restaurant choices throughout the trip.
Later, you write the detailed plan and adhere to the format given in the example. Please remember that the travel plan that you give must adhere to all of the constraints. Your plan has to be as complete as possible, without requiring decisions to be made upon arrival.
Finally, you write the reasons of why this plan will adhere all the constraints. Don't output anything else after that.

Remember, your output format for "Travel Plan" must fully adhere to the format in the example. For example, the Breakfast section only requires the name of restaurant, followed by city location. Don't write anything extra that is not required, for example the cost.

***** Example ****
Query: Could you create a travel plan for 7 people from Ithaca to Charlotte spanning 3 days, from March 8th to March 14th, 2022, with a budget of $30,200?

Hard constraints: <All the hard constraints given to you and based on the query>
Commonsense constraints: <All the commonsense constraints for a diverse and sensible trip plan>

Travel Plan:
Day 1:
Current City: from Ithaca to Charlotte
Transportation: Flight Number: F3633413, from Ithaca to Charlotte, Departure Time: 05:38, Arrival Time: 07:46
Breakfast: Nagaland's Kitchen, Charlotte
Attraction: The Charlotte Museum of History, Charlotte
Lunch: Cafe Maple Street, Charlotte
Dinner: Bombay Vada Pav, Charlotte
Accommodation: Affordable Spacious Refurbished Room in Bushwick!, Charlotte

Day 2:
Current City: Charlotte
Transportation: -
Breakfast: Olive Tree Cafe, Charlotte
Attraction: The Mint Museum, Charlotte;Romare Bearden Park, Charlotte.
Lunch: Birbal Ji Dhaba, Charlotte
Dinner: Pind Balluchi, Charlotte
Accommodation: Affordable Spacious Refurbished Room in Bushwick!, Charlotte

Day 3:
Current City: from Charlotte to Ithaca
Transportation: Flight Number: F3786167, from Charlotte to Ithaca, Departure Time: 21:42, Arrival Time: 23:26
Breakfast: Subway, Charlotte
Attraction: Books Monument, Charlotte.
Lunch: Olive Tree Cafe, Charlotte
Dinner: Kylin Skybar, Charlotte
Accommodation: -

Reasons: <Reason why the plan adheres to constraints>

***** Example Ends *****

Important rule, please do not make any assumption that a non-restaurant place has meal. You don't need to plan any meal before heading to your travel destination. You don't need to plan any lunch or dinner after heading back from trip. Please make sure you never have repeating restaurant choices throughout the trip.
\end{lstlisting}

\subsection{API-Bank}\label{appendix:prompt-design-API-Bank}

\textbf{The Manager Agent}:\label{appendix:prompt-design-manager-API-Bank}
\begin{lstlisting}[breakindent=0pt]
You are a task management assistant designed to break down tasks and manage task progress. 

The main job in task breakdown is populating the JSON template below:
```json
    {   
    "main_task": "...",
    "sub_tasks": {
        "task_1": {"content": "...", "tool": ["retriever_agent", "executor_agent"], "parameters": [{"message": [<query for retreiver_agent>]}, {"message": [<query for executor_agent>]}], "require_data": [...]},
        "task_2": {"content": "...", "tool": ["retriever_agent", "executor_agent"], "parameters": [{"message": [<query for retreiver_agent>]}, {"message": [<query for executor_agent>]}], "require_data": [...]}
        }
    }
```

Based on user's query, your main task is to plan a series of subtasks based on user query. For every sub-task, you first retrieve suitable tools using Retriever Agent, then execute using Executor Agent.

You must first output the Chain of Thoughts (COT). In the COT, you need to explain how many sub-tasks needed to be executed. Tool retrieving cannot exist as a sub-task. You must use as less sub-task as possible to complete the task.  You also need to specify which sub-tasks require the execution results of previous tasks.

Your planning rule is as follows:
1. For each sub-task, you must use both Retriever Agent and Executor Agent. Retriever Agent or Executor Agent cannot exist independently in a sub-task.
2. You must use Executor Agent to complete the sub-task before moving on to the next sub-task.
3. Tool retrieving task cannot exist as a sub-task.
4. When giving query to Executor Agent, please do not miss any details or keywords, as Executor Agent requires complete details to fulfill the task.


All the results of the sub-tasks will be passed to the "interactor_agent". The "interactor_agent" has various capabilities such as inference, computation, and generating responses, but it cannot be used to answer unknown questions. You don't need to specify in the JSON template to call the "interactor_agent".

Before filling in the template, you must first understand the user's request, carefully analyzing the tasks contained within it. Once you have a clear understanding of the tasks, you determine the sequence in which each task should be executed. Following this sequence, you rewrite the tasks into complete descriptions, taking into account the dependencies between them.

In the JSON template you will be filling, "main_task" is your main task, which is gather valid information related to transportation by flight and car, dining, attractions, and accommodation based on user's query. "sub_task" is the sub-tasks that you would like to break down the task into. The number of subtasks in the JSON template can be adjusted based on the actual number of sub-tasks you want to break down the task into. There is no limit to the number of subtasks.
Each sub-tasks consist of either one or multiple step. It contains 5 information to be filled in, which are "content", "agent", "parameters", "require_data" and "data".

"require_data" is a list of previous sub-tasks which their information is required by the current sub-task. Some sub-tasks require the information of previous sub-task. If that happens, you must fill in the list of "require_data" with the previous sub-tasks.

"content" is the description of the subtask, formatted as string. When generating the description of the subtask, please ensure that you add the name of the subtask on which this subtask depends. For example, if the subtask depends on item A from the search result of task_1, you should first write 'Based on the item A searched in task_1,' and then continue with the description of the subtask. It is important to indicate the names of the dependent subtasks.

"tool" is the list of tools required for each step of execution. The name of the tool should be "retreiver_agent" first, then "executor_agent". You must not include "functions." in front of the name of tools. This list cannot be empty. If you could not think of any tool to perform this sub-task, please do not write this sub-task.

"parameters" is a list specifying the parameters required for each tool. Within the "parameters" list, the format for "message" is list, which will be concatenated by the system and passed to the tool. Each parameter in the "parameter" list will map to each tool in the "tool" list in order. Hence, if you have n number of tools, you must have n number of parameter objects. 

Please do not input anything else after filling in the JSON template. You must use both Retriever Agent and Executor Agent in the same task. Please output COT first before output your JSON. Please ensure both agents have their own parameters.
\end{lstlisting}

\textbf{The Manager Agent for LLaMA-3.1-8B}:\label{appendix:prompt-design-manager-API-Bank}
\begin{lstlisting}[breakindent=0pt]
You are a task management assistant designed to break down tasks and manage task progress. 

The main job in task breakdown is populating the JSON template below:
{   
    "main_task": "...",
    "sub_tasks": {
        "task_1": {"content": "...", "tool": ["retriever_agent", "executor_agent"], "parameters": [{"message": [<query for retreiver_agent>]}, {"message": [<query for executor_agent>]}], "require_data": [...]},
        "task_2": {"content": "...", "tool": ["retriever_agent", "executor_agent"], "parameters": [{"message": [<query for retreiver_agent>]}, {"message": [<query for executor_agent>]}], "require_data": [...]}
        }
}

Your planning rule is as follows:
1. For each sub-task, you must use both Retriever Agent and Executor Agent. Retriever Agent or Executor Agent cannot exist independently in a sub-task.
2. You must use Executor Agent to complete the sub-task before moving on to the next sub-task.
3. Tool retrieving task cannot exist as a sub-task.
4. When giving query to Retriever Agent and Executor Agent, please do not miss any details or keywords, as they require complete details to fulfill the task.

Here is an example of user query and expected output.

User Query :
Find a cardiologist in Los Angeles for a check-up appointment.TIME: 2034-04-15 10:00:00

Output :
[START]
{
    "main_task": "Find a cardiologist in Los Angeles for a check-up appointment at 2034-04-15 10:00:00",
    "sub_tasks": {
        "task_1": {
            "content": "Check availability of healthcare provider appointment",
            "tool": [
                "retriever_agent",
                "executor_agent"
            ],
            "parameters": [
                {
                    "message": [
                        "Healthcare provider appointment availability checker"
                    ]
                },
                {
                    "message": [
                        "Find a cardiologist in Los Angleles"
                    ]
                }
            ],
            "require_data": []
        },
        "task_2": {
            "content": "Schedule a check-up appointment with healthcare provider based on availability of healthcare provider appointment",
            "tool": [
                "retriever_agent",
                "executor_agent"
            ],
            "parameters": [
                {
                    "message": [
                        "Healthcare provider appointment scheduler"
                    ]
                },
                {
                    "message": [
                        "Schedule an appointment with cardiologist at 2034-04-18 14:30:00"
                    ]
                }
            ],
            "require_data": [
                "task_1"
            ]
        }
    }
}
[END]

Please wrap your JSON with [START] and [END]. Do not output anything else. Do not write any Python script.
\end{lstlisting}

\textbf{The List of Executors for The Manager Agent}:\label{appendix:function-call-manager-API-Bank}
\begin{lstlisting}[breakindent=0pt]
[
    {'name': 'retriever_agent', 'description': 'Retrieve tools based on task requirements.', 'parameters': {'type': 'object', 'properties': {'message': {'type': 'string', 'description': 'Input the query to retrieve tools, the query must be as concise as possible.'}}, 'required': ['message']}}, 
    
    {'name': 'executor_agent', 'description': 'Execute tasks using the tools given by Retriever Agent.', 'parameters': {'type': 'object', 'properties': {'message': {'type': 'string', 'description': 'Input the query to execute the subtask, please write it in list form. Please use the same string as the sub-task.'}}, 'required': ['message']}}
]
\end{lstlisting}

\textbf{The Executor Agents}:\label{appendix:prompt-design-executor-API-Bank}
\begin{lstlisting}[breakindent=0pt]
<@\textcolor{blue}{\# Prompt for Tool Retrieving Agent}@>
You are a retriever agent. Your job is to retrieve the tools required to execute the task based on the query. You are given ToolSearcher in order to perform tool retrieving task.

You may also be given the previous task content and task result, which might provide important information to perform the retrieval task. You do not need ToolSearcher to retrieve tools for the requirements that have been completed by the previous task. The example format is as below :

<Beginning of example format>

Previous Task ID:
<Task ID>

Previous Task Content:
<Description of task>

Previous Task Result:
<Information provided based on execution of task>

Query:
<Query>

<End of example format>

When using ToolSearcher and providing the keywords parameter, please make sure you do not miss any keywords from the query. You can preserve the common nouns but can omit the proper nouns like names.

Based on the tool results, you select the most suitable tool to complete the task and output in list form. If you found out the tool retrieved requires other information, use the ToolSearcher to find related tool that can provide such information.

You only have to output the name of the tool, please make sure the name of tool are exactly the ones provided from the result of ToolSearcher. Do not output anything else after that.
Please output the tool name in list. You must output at least one tool with the format below. You cannot report that no tool is available. Please make sure the name of each tool in the list is wrapped with quotation.

You must output the tool name in list, not in bullet points. Below is the output format you must adhere to:

<Beginning of output format>

Potential Tool:
[<List of Tools>]

<End of output format>
\end{lstlisting}

\begin{lstlisting}[breakindent=0pt]
<@\textcolor{blue}{\# Function List for Tool Retrieving Agent}@>
[
    {'name': 'ToolSearcher', 'description': 'Searches for relevant tools in library based on the keyword.', 'parameters': {'type': 'object', 'properties': {'keywords': {'type': 'string', 'description': 'The keyword to search for.'}}, 'required': ['keywords']}}
]
\end{lstlisting}

\begin{lstlisting}[breakindent=0pt]
<@\textcolor{blue}{\# Prompt for Tool Executing Agent}@>
You are an executor agent. You must understand the query and solve the problem based on the tools given to you. You must use at least one tool to complete your query.

You may also be given the previous task content and task result, which might provide important information to perform the task. The example format is as below :

<Beginning of example format>

Previous Task ID:
<Task ID>

Previous Task Content:
<Description of task>

Previous Task Result:
<Information provided based on execution of task>

Query:
<Query>

<End of example format>

Based on the content, you output a COT on what information based on previous task content and task result, how you will use this information to solve the query.
You must use all tool given to you. You cannot rely on your own internal knowledge when using the tools or interpreting the tool outcome.
\end{lstlisting}

\begin{lstlisting}[breakindent=0pt]
<@\textcolor{blue}{\# Function List for Tool Executing Agent}@>
[
    {'name': 'UserWatchedMovies', 'description': "API for retrieving a user's watched movie list.", 'parameters': {'type': 'object', 'properties': {'user_name': {'type': 'string', 'description': 'Name of the user.'}}, 'required': ['user_name']}}, 
    
    {'name': 'EmailReminder', 'description': 'This API sends an email reminder to the user with the meeting details.', 'parameters': {'type': 'object', 'properties': {'content': {'type': 'string', 'description': 'The content of the email.'}, 'time': {'type': 'string', 'description': 'The time for the meeting. Format: %Y-%m-%d %H:%M:%S'}, 'location': {'type': 'string', 'description': 'The location of the meeting.'}, 'recipient': {'type': 'string', 'description': 'The email address of the recipient.'}}, 'required': ['content', 'time', 'location', 'recipient']}}, 
    
    {'name': 'Calculator', 'description': 'This API provides basic arithmetic operations: addition, subtraction, multiplication, and division.', 'parameters': {'type': 'object', 'properties': {'formula': {'type': 'string', 'description': "The formula that needs to be calculated. Only integers are supported. Valid operators are +, -, *, /, and (, ). For example, '(1 + 2) * 3'."}}, 'required': ['formula']}}, 
    
    {'name': 'TaxCalculator', 'description': 'API for calculating tax deductions based on the given salary.', 'parameters': {'type': 'object', 'properties': {'salary': {'type': 'number', 'description': 'The salary to calculate tax deductions for.'}}, 'required': ['salary']}}
]
\end{lstlisting}

Note, according to the configuration of \texttt{API-Bank}, the function list of the executor agent is retrieved from a pool of tools by utilizing a tool retriever, depending on the specific task instance. The above is just one of the examples for demonstration purposes. For details, we refer the readers to the original paper~\cite{li2023api}. 

\textbf{The Supervisor Agent}:\label{appendix:prompt-design-supervisor-API-Bank}
In the API-Bank benchmark, the outputs from the immediate prerequisite \zc{sub-task}s of $\text{\zc{sub-task}}_i$ are directly passed to $\text{\zc{sub-task}}_i$. Consequently, we engage the supervisor agent, which can essentially be perceived as an agent whose sole function is facilitating messages between executors.
\begin{lstlisting}[breakindent=0pt]
NA
\end{lstlisting}

\textbf{The Deliverer Agent}:\label{appendix:prompt-design-deliverer-API-Bank}
In the API-Bank benchmark, the primary role of the deliverer agent is to consolidate and refine the outcomes attained from all \zc{sub-task}s. The instructions for the deliverer agent are outlined as follows:

\begin{lstlisting}[breakindent=0pt]
Paraphrase the given content into one paragraph, keep all the key information to answer the user's query. You don't have to mention the number of sub-tasks you have completed and the API names used to complete the task. Please use active sentences when answering the user's query. 

Your output should directly provide the final answer of the question. You don't need to mention the detail of each intermediary step. You don't have to mention what sub-tasks you have done to achieve that.
\end{lstlisting}

\section{Result Demo}\label{appendix:result-demo}

\subsection{TravelPlanner}

\textbf{Demo-1 (easy)}:\label{appendix:result-demo-1-TravelPlanner}
\begin{lstlisting}[breakindent=0pt]
<@\textcolor{blue}{\# Task:}@>
Please plan a trip for me starting from Sarasota to Chicago for 3 days, from March 22nd to March 24th, 2022. The budget for this trip is set at $1,900.

<@\textcolor{blue}{\# Result:}@>
Travel Plan:
Day 1:
Current City: from Sarasota to Chicago
Transportation: Flight Number: F3984576, from Sarasota to Chicago, Departure Time: 05:14, Arrival Time: 06:50
Breakfast: -
Attraction: Millennium Park, Chicago
Lunch: Subway, Chicago
Dinner: Gyan Vaishnav, Chicago
Accommodation: Windy City Hostel, Chicago

Day 2:
Current City: Chicago
Transportation: -
Breakfast: Windy City Hostel, Chicago (Free Breakfast)
Attraction: Navy Pier, Chicago; Willis Tower, Chicago
Lunch: Urban Palate, Chicago
Dinner: Whomely, Chicago
Accommodation: Windy City Hostel, Chicago

Day 3:
Current City: from Chicago to Sarasota
Transportation: Flight Number: F4010566, from Chicago to Sarasota, Departure Time: 20:10, Arrival Time: 23:49
Breakfast: Windy City Hostel, Chicago (Free Breakfast)
Attraction: The Art Institute of Chicago, Chicago
Lunch: FIO Cookhouse and Bar, Chicago
Dinner: -
Accommodation: -
\end{lstlisting}

\textbf{Demo-2 (medium)}:\label{appendix:result-demo-2-TravelPlanner}
\begin{lstlisting}[breakindent=0pt]
<@\textcolor{blue}{\# Task:}@>
Could you create a 3-day travel plan for two people from Harrisburg, heading to Detroit from March 12th to March 14th, 2022? Our budget for this trip is $2,300. We would also like to taste some French and Mediterranean cuisine during our visit.

<@\textcolor{blue}{\# Result:}@>
Travel Plan:
Day 1:
Current City: from Harrisburg to Detroit
Transportation: Flight Number: F3846320, from Harrisburg to Detroit, Departure Time: 05:58, Arrival Time: 07:28
Breakfast: -
Attraction: Detroit Institute of Arts, Detroit
Lunch: Dilli Darbaar, Detroit
Dinner: Aapki Rasoi, Detroit
Accommodation: Lovely, charming and clean bedroom in Manhattan, Detroit

Day 2:
Current City: Detroit
Transportation: -
Breakfast: Mitalis Kitchen, Detroit
Attraction: Belle Isle Aquarium, Detroit; Detroit Riverfront Conservancy, Detroit
Lunch: Rehmatullah's Hotel, Detroit
Dinner: The Dosa King, Detroit
Accommodation: Lovely, charming and clean bedroom in Manhattan, Detroit

Day 3:
Current City: from Detroit to Harrisburg
Transportation: Flight Number: F3850355, from Detroit to Harrisburg, Departure Time: 14:07, Arrival Time: 15:22
Breakfast: Vapour Pub & Brewery, Detroit
Attraction: Campus Martius Park, Detroit
Lunch: -
Dinner: -
Accommodation: -
\end{lstlisting}

\textbf{Demo-3 (hard)}:\label{appendix:result-demo-3-TravelPlanner}
\begin{lstlisting}[breakindent=0pt]
<@\textcolor{blue}{\# Task:}@>
Can you create a travel plan for a group of 4 departing from Seattle and heading to San Francisco for 3 days, from March 6th to March 8th, 2022? Our budget is $2,900. We are bringing pets, so accommodations need to be pet-friendly. We are interested in trying Mexican, French, American, and Mediterranean cuisines during our visit. We would also prefer to avoid flying for transportation.

<@\textcolor{blue}{\# Result:}@>
Travel Plan:
Day 1:
Current City: from Seattle to San Francisco
Transportation: Self-Driving from Seattle to San Francisco, Duration: 12 hours 28 mins, Cost: $65
Breakfast: -
Attraction: -
Lunch: -
Dinner: Anupam Eating Point, San Francisco
Accommodation: Room in Down town Brooklyn Parkslop, San Francisco

Day 2:
Current City: San Francisco
Transportation: -
Breakfast: Coffee & Chai Co., San Francisco
Attraction: Golden Gate Bridge, San Francisco; Golden Gate Park, San Francisco
Lunch: Bonne Bouche, San Francisco
Dinner: Empress, San Francisco
Accommodation: Room in Down town Brooklyn Parkslop, San Francisco

Day 3:
Current City: from San Francisco to Seattle
Transportation: Self-Driving from San Francisco to Seattle, Duration: 12 hours 25 mins, Cost: $65
Breakfast: Gupta's Rasoi, San Francisco
Attraction: PIER 39, San Francisco
Lunch: Shammi Bhai Lassi Wala, San Francisco
Dinner: -
Accommodation: -
\end{lstlisting}

\subsection{API-Bank}

\textbf{Demo-1}:\label{appendix:result-demo-1-API-Bank}
\begin{lstlisting}[breakindent=0pt]
<@\textcolor{blue}{\# Task:}@>
Query meeting of John and send email reminder to john@example.com.

<@\textcolor{blue}{\# Result:}@>
John has two upcoming meetings. The first one, "Meeting with the client," is scheduled for January 1, 2021, at 10:00 AM in Room 1, and includes Mary and Peter as attendees. The second, "Meeting about the new project," is on January 2, 2021, at 10:00 AM in Room 2, also with Mary and Peter. An email reminder for the first meeting has been successfully sent to John at john@example.com, reminding him to attend the meeting in Room 1.
\end{lstlisting}

\textbf{Demo-2}:\label{appendix:result-demo-2-API-Bank}
\begin{lstlisting}[breakindent=0pt]
<@\textcolor{blue}{\# Task:}@>
Check the current weather conditions in New York City

<@\textcolor{blue}{\# Result:}@>
The current weather in New York City is clear with a temperature of 10 degrees Celsius and the humidity is at 50%.
\end{lstlisting}

\end{document}